\documentclass[fleqn,10pt]{wlscirep}
\usepackage[utf8]{inputenc}
\usepackage[T1]{fontenc}
\title{Prediction of Transportation Index for Urban Patterns in Small and Medium-sized Indian Cities using Hybrid RidgeGAN Model}

\author[1]{Rahisha Thottolil}
\author[1]{Uttam Kumar}
\author[2,1,3*]{Tanujit Chakraborty}
\affil[1]{Spatial Computing Laboratory, Center for Data Sciences, International Institute of Information Technology, Bangalore, 560100, India.}
\affil[3]{Department of Science and Engineering, Sorbonne University Abu Dhabi, United Arab Emirates.}
\affil[3]{School of Business, Woxsen University, Telengana, India.}

\affil[*]{tanujit.chakraborty@sorbonne.ae}


\begin{abstract}

The rapid urbanization trend in most developing countries including India is creating a plethora of civic concerns such as loss of green space, degradation of environmental health, clean water availability, air pollution, traffic congestion leading to delays in vehicular transportation, etc. Transportation and network modeling through transportation indices have been widely used to understand transportation problems in the recent past. This necessitates predicting transportation indices to facilitate sustainable urban planning and traffic management. Recent advancements in deep learning research, in particular, Generative Adversarial Networks (GANs), and their modifications in spatial data analysis such as CityGAN, Conditional GAN, and MetroGAN have enabled urban planners to simulate hyper-realistic urban patterns. These synthetic urban universes mimic global urban patterns and evaluating their landscape structures through spatial pattern analysis can aid in comprehending landscape dynamics, thereby enhancing sustainable urban planning. This research addresses several challenges in predicting the urban transportation index for small and medium-sized Indian cities. A hybrid framework based on Kernel Ridge Regression (KRR) and CityGAN is introduced to predict transportation index using spatial indicators of human settlement patterns. This paper establishes a relationship between the transportation index and human settlement indicators and models it using KRR for the selected 503 Indian cities. This nonlinear KRR model helps in deriving the transportation network for GAN-generated human settlements through the settlement indicators. Our approach leverages human settlement indices, which capture information about demographics and urban land use to predict the transportation index. The proposed hybrid pipeline, we call it RidgeGAN model, can evaluate the sustainability of urban sprawl associated with infrastructure development and transportation systems in sprawling cities. Experimental results show that the two-step pipeline approach outperforms existing benchmarks based on spatial and statistical measures. By predicting future urban patterns, this study can help in the creation of more livable and sustainable cities, particularly by improving transportation infrastructure in small and medium-sized Indian cities.

\end{abstract}
\begin{document}

\flushbottom
\maketitle

\thispagestyle{empty}

% \noindent Please note: Abbreviations should be introduced at the first mention in the main text – no abbreviations lists. Suggested structure of main text (not enforced) is provided below.

\section*{Introduction}

Mapping urban land use dynamics has been valuable research in urban studies over several decades. The advancement in remote sensing technology makes it possible to track the spatiotemporal changes in urban landscape structures with relatively high accuracy and on a required scale \cite{lv2021land}. The spatial distribution of land use activities (residential, commercial, industrial, etc.) including the transportation system is important for understanding the current urban centers and for planning future city development \cite{li2021understanding}. Land use and land cover maps become a starting point for modeling urban patterns and they can infer the future urban growth and the direction of land expansion of cities to inform urban planners and government policymakers towards sustainable urban planning \cite{mahmoudzadeh2022urban,rana2021prediction, vani2020assessment}. While urban areas continue to experience rapid growth, they pose new challenges to the nation, especially for developing and underdeveloped countries \cite{sanga2022top}. Hence, urban growth prediction models and related studies have  become a hot topic that has been extensively and deeply investigated. The existing urban growth models %(eg. CA, MC-AHP, SLEUTH, agent-based, and PLUS)
\cite{falah2020urban, mathioulakis2017using, hamdy2016applying, liang2021understanding} include the driving factors which affect urban expansion. These factors influencing urban expansion include population growth, economic development, urbanization, transportation infrastructure, topography, and land use regulations \cite{pravitasari2018identifying}. However, in developing and underdeveloped nations, where urban expansion is more likely to occur, data on driving forces are hard to obtain and are often expensive to collect.

According to the UN report \cite{united2022world}, India is the most populous country in the world. From 1901 to 2011, the country's urban population expanded by around 14 times \cite{ritchie2018urbanization}. Although largely unequal, its increase is not skewed and is not limited to one region across the nation. The skyrocketing living costs in metropolitan areas and increasing house rents discourage enterprises from investing in major cities. Therefore, it is essential to assess the settlement pattern and infrastructure facilities of small and medium towns as an alternate option to larger metropolitan cities \cite{ganguly1997integrated}. These towns sometimes called the ``next billion'' markets, will be important in propelling the expansion of the national economy in India \cite{ganguly201614}. Further to the intricacy of urban settlement, more basic infrastructure and facilities are required for these regions. 

Various reports in recent years have estimated a massive demand for funding urban infrastructures in developing countries. The World Bank estimates that nearly 70000 Billion (INR) of investment in urban India will be required to meet the growing population demands in the next 15 years until 2036 (in 2020 prices) \cite{athar2022financing}. For example, the Indian Government introduced a scheme called the Integrated Development of Small and Medium Towns (IDSMT) project that aims to encourage the planned and sustainable growth of the nation's small and medium-sized towns or cities \cite{ganguly201614}. The Ministry of Urban Development, Government of India introduced this scheme in 2005, and it offers financial and technical assistance to local governments in order to help them develop their towns’ infrastructure and fundamental services. This scheme focuses on the expansion of small and medium-sized towns (the population size is up to 500K), which may act as growth hubs for the nearby rural regions, in order to promote inclusive growth and balanced regional development. By providing funding for the construction of fundamental utilities including water supply, sanitation, solid waste management, and urban transportation, the program seeks to solve the infrastructure deficit and service shortages in these communities. The overall goal of the IDSMT program is to support sustainable urban growth and raise the standard of living in India's small and medium-sized towns. As a result, it is important to examine the small and medium towns (Tier 3 and above cities with populations up to 500K) in India. Adopting cutting-edge technologies can have a significant impact on enhancing Government's effectiveness in improving planning and decision-making, problem-solving, accelerating development, and deployment \cite{engstrom2020government}. 

To mitigate this urban planning challenge, recent developments in machine learning and deep learning have become handy tools for urban planners and geoscience practitioners. Deep learning models such as Generative Adversarial Networks (GAN) can approximate complex, high-dimensional probability distributions \cite{Goodfellow2014GenerativeAN}. GANs have achieved numerous state-of-the-art breakthroughs in the fields of computer vision \cite{wang2021generative}, natural language processing \cite{pan2019recent}, and more recently in urban science and geospatial domain \cite{zhu2020spatial, Wu2022GenerativeAN}. % As a result, a few advanced deep neural network methods generative models, in particular, 
Several GAN-based models have been proposed to simulate hyper-realistic urban land use maps and generate synthetic urban universe without considering the driving factors, see for example CityGAN \cite{Albert2018ModelingUP} and MetroGAN \cite{zhang2022metrogan}. Among these, CityGAN \cite{Albert2018ModelingUP} simulates urban patterns using global urban land-use inventory and builds an ``urban universe'' to reproduce the complex spatial patterns observed in global cities. An extension to CityGAN by incorporating geographical knowledge is called Metropolitan GAN (MetroGAN) \cite{zhang2022metrogan} which learns hierarchical features for urban morphology simulation. Another deep learning method, namely U-Net \cite{ronneberger2015unet}, is also applied to generate future urban cities using water bodies, digital elevation models, and nighttime lights as inputs \cite{Wu2022GenerativeAN}. The application of previous GAN-based urban models was limited to the generation of urban patterns. There are limited works on quantifying the urban pattern of GAN-generated images and predicting the transportation metric (representing urban infrastructure) for these new urban regions. Thus, quantification and modeling of landscape patterns of those GAN-simulated cities remain an unattempted problem. The structure of a landscape emerges from the characteristics of the individual elements of an ecosystem and their spatial configuration \cite{forman1981patches, forman1986landscape, gokyer2013understanding, mobaied2016importance}. Human Settlement Indices (HSI), e.g., Class Area (CA), Number of Patches (NP), Largest Patch Index (LPI), Clumpiness Index (CLUMPY), Aggregation Index (AI), and Normalized Landscape Shape Index (NLSI) \cite{Aithal2020UrbanGP, Sudhira2012EffectOL, wilson2003development} provide some concrete information about the landscape structures and therefore contributes in the prediction of Transportation Index (TI). This paper makes an attempt to answer the following challenging questions: 
\begin{enumerate}[label = (\alph*)]
    \item How to generate an urban universe for India based on spatial patterns via learning urban morphology?
    \item Is there a relationship between HSI and TI in small and medium cities in India?
    \item How to predict (forecast) TI for synthetic urban cities generated by CityGAN for developing countries like India? 
\end{enumerate}

To generate small and medium-sized Indian cities with CityGAN, we collected the World Settlement Footprints (WSF 2019) maps which are publicly available and the best representations of urban patterns as input features \cite{esch2017breaking, esch2018global}. Then, we build a city image database of 503 small and medium Indian cities whose populations range between 20K and 500K. Each city image represents $10.5 \times 10.5$ km covering the urban center and surrounding regions. We also explored existing spatial and statistical measures to evaluate the performance of CityGAN in Indian cities due to the complex nature of individual cities with varying structural and hierarchical properties \cite{Albert2018ModelingUP}. Assessing the spatial relationship between urban patterns (human settlement) and the transportation index of actual cities can help to build a model for predicting the transportation index for generated cities. We used different linear and nonlinear measures of statistical correlations to establish this spatial relationship. Furthermore, we propose a hybrid model (namely, RidgeGAN) to predict the transportation index for simulated urban patterns. Fig. \ref{fig:methods} depicts the methodological framework. In RidgeGAN, a supervised learning model (KRR) builds a relationship between the human settlement patterns and the characteristics of the urban road transportation system and implements this to predict TI for the GAN-simulated urban universe for India. Our proposal has numerous applications, ranging from understanding urban land patterns to predicting relevant urban infrastructure facilities to guiding policymakers toward a better and more inclusive planning process. %\cite{zhao2021comparison}

\section*{Background and related work}

\subsection*{Applications of GANs in geospatial field}

Deep learning has reached a significant milestone in geospatial research, computer vision and other cutting-edge technologies \cite{roberts2022principles}. GAN \cite{Goodfellow2014GenerativeAN}, an essential subfield of unsupervised deep learning, has opened a new vista for geoscience research in recent years. GANs are utilized to generate data that is close to a given training set which can be images, texts and tabular data \cite{Hong2017HowGA}. Geoscientists and urban planners have adopted this new deep learning methodology for handling geophysical and remote sensing data. In remote sensing, MARTA GANs were proposed for producing fake satellite images of urban environments \cite{Lin2016MARTAGU}. It consists of a discriminator network that receives both real and synthetic images as input and predicts whether each image is genuine or synthetic, as well as a generation network that uses random noise as input to create synthetic images. Further, Spatial Generative Adversarial Networks (SpaGANs) \cite{Jetchev2016TextureSW} were introduced for synthesizing textures by incorporating spatial information (such as the position and orientation of the texture) into the generator and discriminator networks. A comprehensive review of GANs demonstrates promising performance in the built environment, from processing large-scale urban mobility data and remote sensing images at the regional level to performance analysis and design generation at the building level \cite{Wu2022GenerativeAN}. 

In addition, GANs were also applied to modeling global urban patterns. For example, CityGANs \cite{Albert2018ModelingUP}, conditional GANs \cite{albert2019spatial}, and MetroGANs \cite{zhang2022metrogan} was built to generate urban land patterns by training the generator networks to generate synthetic images of urban areas that closely resemble real urban areas and it will be useful for analyzing urban human settlement data from space-based sensors. For more accurate urbanization parameters or spatial indices estimates in locations where local data is unavailable or impossible to collect, these models are very effective in simulating urban land use patterns. GANs are used to model hyper-realistic settlement patterns since they do not make any assumptions about the data distribution and can generate real-like samples from the latent space in a straightforward manner. This unique property lends GANs to a variety of geospatial applications, including image synthesis, image attribute editing, image translation, domain adaptation, and other computing fields \cite{Xing2022UnsupervisedDA}. 

\subsection*{Transportation and urban landscape structures}

%\subsubsection*{Landscape Metrics} 

The spatial structure of a city is extremely complex and is constantly evolving. Therefore, there are significant attempts to analyze cities, and thus to link urban policy to shape cities. Delineating homogeneous/heterogeneous human settlements, quantifying them, and analyzing their diversity and spatial organization are necessary to assess their structures and spatial patterns. Due to this, urban researchers utilized landscape metrics to quantify the qualities of the landscape related to shape, pattern, and area by measuring the structure and spatial distribution of settlements. Landscape metrics were originally introduced in ecological studies that reflected social, cultural, and ecological richness and heterogeneity \cite{Wu2002KeyIA}. Also, progressive and well-functioning urban planning departments can use spatial indicators to regularly monitor urban development and, when necessary, propose regulatory or public investment action \cite{bertaud2003spatial}. These indicators can also evaluate the geometrical characteristics of ecological processes and landscape elements, as well as their relative locations and distribution \cite{Brown2012SocialLM}. The effect of landscape metrics on spatial patterns was studied to quantify landscape structures and these metrics can statistically determine the outcome \cite{Sudhira2012EffectOL}. Spatial patterns of urban growth and landscape metrics were studied for various cities in India \cite{bhat2015spatial, Aithal2020UrbanGP}. The integration of efficient urban structures and comprehensive transportation metrics plays a vital role in fostering sustainable development and improving the overall livability of cities.

%\subsubsection*{Transportation Metrics}
Understanding the interaction between transportation infrastructures and urban pattern areas is always critical for driving smooth urban services. The investigation into the development of mathematical models for studying the relationship between transportation metrics and urban land use began in the early 1960s and technological advancements brought us to an era of integrated land use transportation modeling \cite{Zhong2022AdvancesII}. Several road network models have been developed to solve transportation problems. Most of the existing transportation models for prediction now in use are based on simple linear regression models. An inverse relationship between urban growth and transportation was found for the middle east regions \cite{Aljoufie2011UrbanGA}. Their analysis suggested that urban population growth has increased urban trips and increased travel demand due to transportation infrastructure.

In a recent study, the link between population and the characteristics of the road network in the Lebanese Republic was investigated using a multivariate regression model to estimate the population count based on various data sources and statistical modeling techniques \cite{Allaw2019ARS}. However, linear regression models have the drawback of excluding all variables which are not linearly connected, and multicollinear variables adversely affect the model. Besides, it is proved that the relationship between transportation features and their influencing factors is not always linear in nature. Predicting the transportation index is an important step towards minimizing traffic congestion and providing critical information to individual travelers as well as various Government sectors to plan the city in a sustainable way. A support vector regression (SVR) approach was used to predict traffic flow from California highways using different types of kernels \cite{zeroual2021predicting}. A road network density (one of the transportation indices) prediction model was proposed using highway capacity, and turning probabilities manual methods were used to determine the shortest cycling time in metropolitan areas \cite{Budiarto2014RoadDP}. The model could aid in vehicle distribution and congestion relief in urban areas. Further, the concept of graph theory was used to analyze the topology of road networks in an Indian city to better understand the connectivity and coverage of the existing road transportation system \cite{Thottolil2021AssessmentOT}. Their findings indicate that there is a strong relationship between road connectivity and coverage and that improving the road network is essential for a reliable and safe road transportation system. To accurately estimate the connectivity index, the paper \cite{Thottolil2021AssessmentOT} proposed a model based on the relationship between the Eta index and Network Density (ND), Edge Graph Density (EGD) and Nodal Graph Density (NGD).

\section*{Results}

The scholarly literature on urban challenges primarily focuses on megacities and large urban centers, but there are a great number of small and medium-sized cities in developing countries that should be prioritized. There is a pressing need to address the challenges of developing transportation networks and settlement patterns in these cities. Previous literature on urban studies does not adequately address the transportation, unplanned city growth and socioeconomic and environmental challenges of small and medium-sized towns \cite{sharma2021small}. In this study, we utilize global training WSF data across India, and we demonstrate a simple and unconstrained GAN model to generate realistic settlement patterns that encompass the diversity of urban forms. We are primarily interested in how small and medium-sized cities are simulated using unsupervised CityGANs. Subsequently, an effective data-driven hybrid model is developed to predict road network density for a given urban settlement pattern.

\subsection*{Study area} 

\begin{figure}[ht]
\centering
\includegraphics[width=\linewidth]{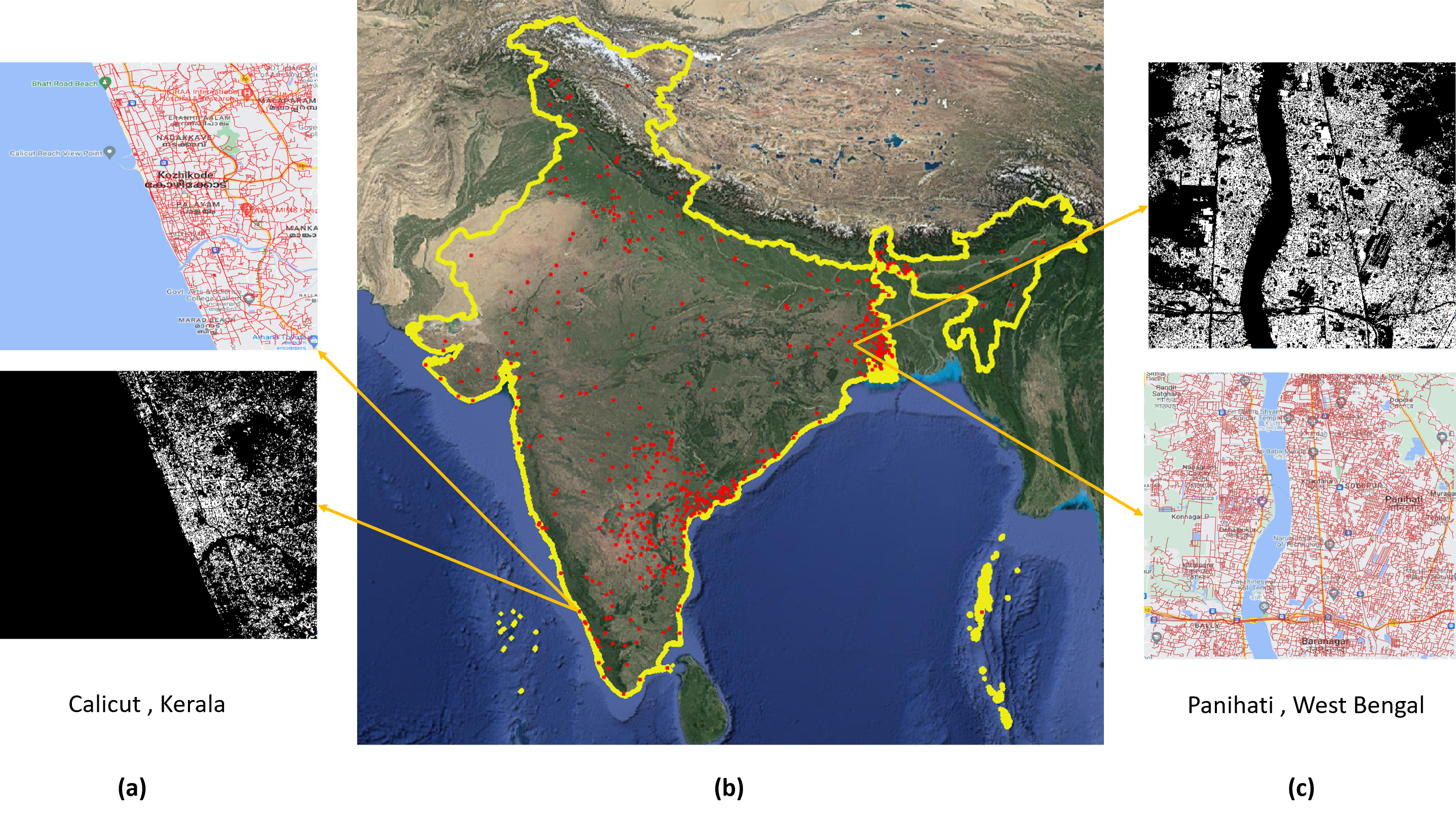}
\caption{(b) Geographical distribution of 503 small and medium-sized Indian cities included in the study (red colored square grids indicate the selected cities); (a) and (c) are examples of human settlement and transportation maps of two random cities, namely Calicut from the state of Kerala and Panihati from the state of West Bengal.}
\label{fig:Studyare_Nature}
\end{figure}

Our study focuses on small and medium-sized Indian cities (South Asia), one of the world's fastest-urbanizing regions. The selection of the study area involved the identification of the geographical location and corresponding demographic data for which the population data from the World Cities database (~\url{https://simplemaps.com/data/world-cities}) were used. Approximately, 503 cities out of 1600 were selected for our study where the population size ranges between 20k to 500k. The geographical location of study areas marked in the map of India is depicted in Fig. \ref{fig:Studyare_Nature}(b) showing visualization of human settlement patterns and corresponding transportation networks (also see Fig. \ref{fig:Studyare_Nature}(a) and  \ref{fig:Studyare_Nature}(c)).

\subsection*{Data collection and preprocessing}
Settlement footprints and transportation network datasets of all the selected cities were collated from various sources over 2019. Settlement maps were procured from the global inventory built-up database called WSF, published by the German Aerospace Center (DLR). These are binary maps (urban area pixels have a value of 1 and non-urban area pixels have a value of 0) derived from multi-temporal space-borne satellites namely Sentinel-1 and Sentinel-2 data that aided in estimating the human settlement pattern of an urban area ($10.5 \times 10.5$  km) with a resolution of around 10m/px. Open source GIS software (QGIS) was used to pre-process (rectify, project, and crop) the images and build a city database. To measure Human Settlement Indices (HSI), existing landscape metrics were selected and computed using Fragstat software \cite{McGarigal1995FRAGSTATSSP}. It includes the packages to compute popular landscape metrics using spatial pattern analysis. We use six settlement indices: Total class area (CA), number of patches (NP), largest patch index (LPI), clumpiness index (CLUMPY), aggregation index (AI) and normalized landscape shape index (NLSI) to estimate the characteristics of human settlement \cite{Aithal2020UrbanGP, Sudhira2012EffectOL}. 

CA is a useful metric to depict the spatial extent of the settlements. It is a composition that specifies the extent of landscape that is made up of a specific class type (e.g., built-up area). The total class area, which is the sum of the areas (${m}^2 $) of all the patches of the relevant patch type divided by 10,000 (converted to hectares) and CA $> 0$ indefinitely. NP in each landscape indicates the degree of fragmentation that counts the number of human settlements or urban patches. The higher the value of NP, the higher is the fragmentation with no limit. At the class level, LPI estimates the percentage of the total landscape area occupied by the largest patch as an indicator of dominance. LPI is calculated by dividing the area (${m}^2 $) of the largest patch of the relevant patch type by the entire landscape area (${m}^2 $) and multiplying the result by 100 (converted to a percentage), i.e., LPI is the percentage of the landscape comprised by the largest patch. LPI values (0 < LPI < 100) decrease from the city center to the outskirts. Another human settlement metric CLUMPY deals with aggregation and disaggregation for adjacent settlements. It shows the frequency with which different pairs of patch types appear side by side. The value ranges from -1 to 1; -1 indicates maximally disaggregated patch type, 0 when the patch type is randomly distributed, and 1 when the patch type is maximally aggregated \cite{McGarigal1995FRAGSTATSSP}. Another metric AI is calculated using an adjacency matrix, which indicates how frequently distinct pairs of patch types (including adjacencies between the same patch type) appear on the map side by side in the settlement map. Its values range from 0 to 100; AI values are less indicating maximum disaggregation and the high AI shows the maximally aggregated single and compact patch. Finally, NLSI provides a measure of class aggregation for which the values ranges from 0 to 1, where 0 means the landscape consists of a single square or maximally compact (i.e., almost square) patch. NLSI increases as the patch type becomes increasingly disaggregated and reaches 1 when the patch type is maximally disaggregated \cite{Aithal2020UrbanGP}.

\begin{figure}[ht]
\centering
\includegraphics[width=\linewidth]{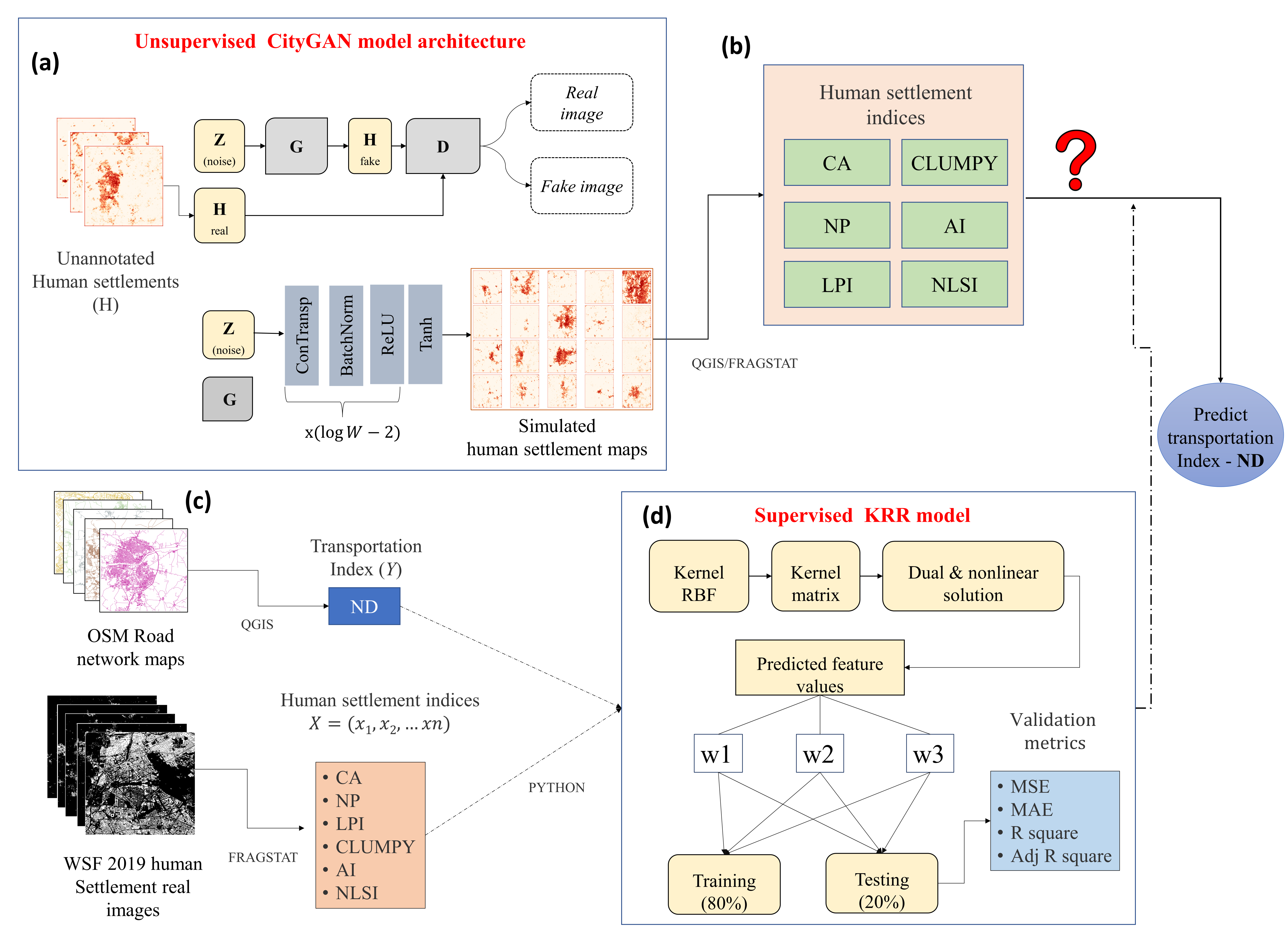}
\caption{Prediction framework of the proposed hybrid RidgeGAN model: (a) Implementing an unsupervised learning model (CityGAN) to generate small and medium-sized Indian cities; (b) Landscape structures of generated cities are measured in terms of human settlement indices (HSI) using spatial landscape metrics; (c) Characteristics of the road network and landscape structures of real cities are measured in terms of HSI and transportation index (TI); (d) Assessing the relations between the settlement patterns and transportation system and building a supervised learning model to predict the transportation index for GAN-generated urban universe.}
\label{fig:methods}
\end{figure}

To compute the Transportation Index (TI), a layer of the road network that has been topologically cleaned and converted into polylines is a prerequisite. The application software used here is integrated with QGIS 3.30 for this purpose. Individual cities and their corresponding road networks were extracted for assessing the spatial patterns of the road systems corresponding to the respective cities. As transportation measures need to be calculated in a metric system, a projected coordinate system was used. Instead of common WGS84 - EPSG:4326, which uses degrees as a unit for distance, the Coordinate Reference System WGS84/UTM-EPSG was used here, to measure road length in meters. All categories of roads such as National highways, State highways, major roads, street roads, residential paths, footways, and service roads were included in this study. To measure the development of the urban road network, the network density of the respective cities was computed as follows:
\begin{equation*}  
\text{Network Density} \; (ND) = \frac{L}{A} = \frac{\text{Total length of the road network}}{\text{City area}};
%\label{eqn:ND} 
\end{equation*} 
where $L$ is determined from road maps and it has been calculated using an open field calculator in QGIS software. Network length specifies the total span of the road network and network density is measured according to the area occupied by road networks (city area), denoted as $A$. Fig. \ref{fig:methods} shows the overall workflow of the proposed hybrid framework used in this study to predict the transportation index for any kind of urban pattern in Indian cities.

\subsection*{Validation metrics}
\begin{figure}[!t]
\centering
\includegraphics[width=5in]{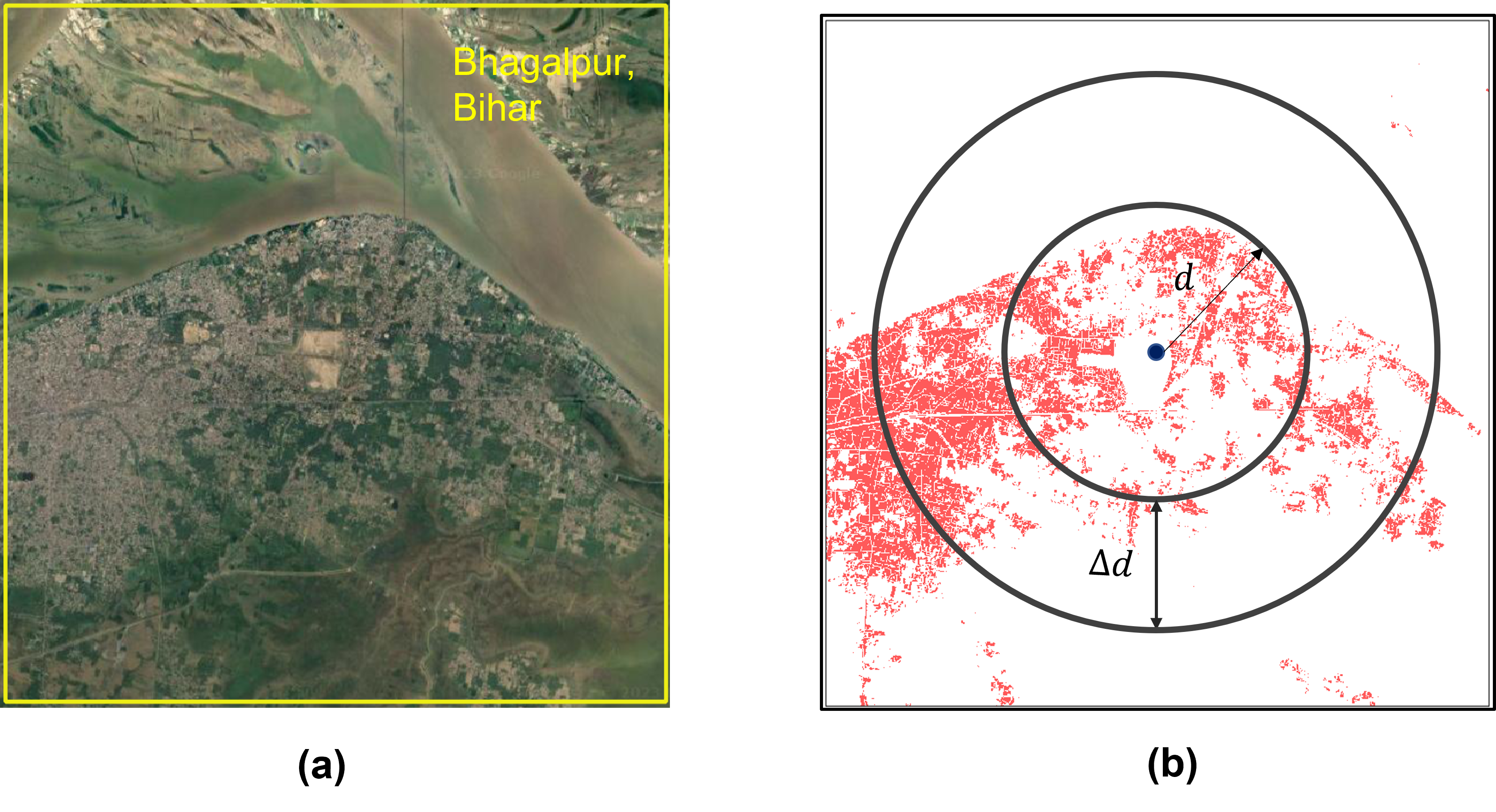}
\caption{(a) Google satellite view of Bhagalpur city (randomly selected city to explain) in Bihar state (b) human settlement map with the method of computing their average radial profiles.}
\label{fig:Bihar}
\end{figure}

In the subsequent section, we estimate the Average Radial Profile (ARP) for real and generated urban patterns to assess the accuracy of CityGAN quantitatively. An example of computing the ARP of a city is illustrated in Fig. \ref{fig:Bihar} and using a peak search algorithm, we determined the polycentricity of actual and simulated scenes from the radial profiles. ARP ($h(d)$ or $x(d)$) represents how much the human settlement area changes as we go out from the city center. As indicated in Fig. \ref{fig:Bihar} (b), we draw rings at a distance of $d$ from the center and a width $\Delta$ of $d$. By averaging the entire settlement area inside rings of width $\Delta d$ at a distance $d$ from the city center, we can compute $ARP$ ($h(d)$). The region inside the ring with radius $d$ is denoted as $R(d)$ and each pixel inside the ring has some build-up area, that is the amount of urbanized area denoted as $H(u,v)$, where $(u,v)$ is a point inside the ring  $H(u,v) \in R(d)$, with
\begin{equation}
\begin{split}
    R(d) \equiv(u, v) \mid(u-u_0) ^2+(v-v_0)^2>d^2 \; \; \text{and} \; \; (u-u_0)^2+(v-v_0)^2 \leq d^2;
\label{eqn:r(d)}
\end{split}
\end{equation}
\begin{equation}
    h(d)=\frac{1}{|R(d)|} \sum_{(u, v) \in R(d)} {h(u,v)};
\label{eqn:h(d)}
\end{equation}
where $R(d)$ can be defined as the collection of all the two-dimensional points included within the ring, $|R(d)|$ indicates the size of set $R(d)$, and $h(d)$ is the average radial profile of a city. 

There are several statistical measures that are used to assess the supervised regression models, for e.g., mean squared error (MSE), mean absolute error (MAE), R-squared ($R^2$), and Adjusted R-squared (Adj $R^2$). MSE is the average squared difference between the predicted and actual values. It is a widely used metric that measures the quality of a regression model whereas MAE is the average absolute difference between the predicted and actual values. It is a robust metric that is less sensitive to outliers than MSE. R-squared measures the proportion of variance in the target variable that is explained by the model. It ranges from 0 to 1, with higher values indicating a higher correlation. Adjusted R-squared is a modified version of R-squared that takes into account the number of predictors in the model. It penalizes the model for adding unnecessary variables and is a better measure of a model's goodness of fit \cite{chakraborty2019hybrid}.

\subsection*{Generating human settlement area from WSF 2019}
\begin{figure}[ht]
\centering
\includegraphics[width=\linewidth]{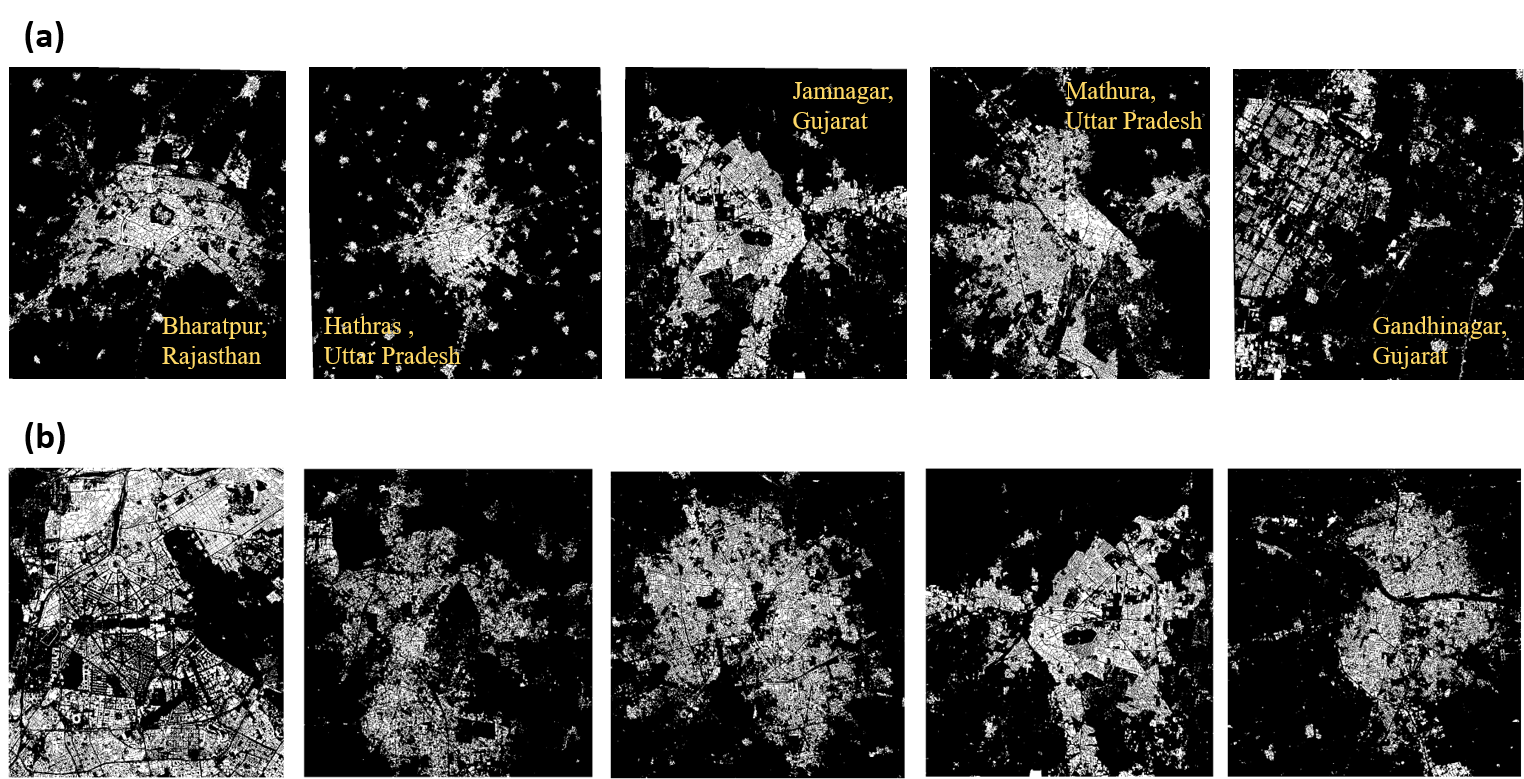}
\caption{Comparison of real urban built land use maps (a) and synthetic maps (b) generated by a CityGAN. The pixel values in each case are in the range [0, 1], where 1 represents the portion of land occupied by buildings. Names of the cities are reported in (a) using yellow color text.}
\label{fig:Fake_real}
\end{figure}

A dataset of real-time human settlement images was collected and pre-processed before training the GAN. We utilized squared shape settlement maps of cities from the WSF 2019 database. We clipped them representing 10.5km × 10.5km spatial extent and resized each image to 256× 256 (43m/px) for optimization purposes and to avoid overfitting. The final input dataset contains 503 binary images and can be formulated as $H_i = {h_1,….,h_n}$, with $H \in R^{W X W}$ and $W=256$ and $n=503$. The source of $H_i$ is an urban binary map (1 and 0 represent urban and non-urban areas respectively). The generator network is trained to generate synthetic urban settlement images by generating random noise and transforming it into an urban image whose distribution matches the real urban images. The discriminator network is trained to distinguish between real and synthetic binary images. Here, generator $G$ takes a random noise vector $Z_{noise}$ as input, which deterministically changes (e.g., by passing it through successive deconvolutional layers if $G$ is a deep CNN) to generate a sample fake human settlement image $H_{fake} = G(z)$. Then the discriminator ($D$) accepts an input image $H$ (which can be an actual human settlement image ($H_{real}$) from an empirical dataset or generated image ($H_{fake}$) synthetically by a generator and outputs the source probability that $H$ is either sampled from the real distribution ($H_{real}$) or produced by generator ($H_{fake}$). Having trained a generator ($G$) (refer Fig. \ref{fig:methods} (a)), we generated synthetic Indian urban settlement patterns of 500 binary images using the CityGAN model. Fig. \ref{fig:Fake_real} illustrates randomly selected real cities (Fig. \ref{fig:Fake_real}(a)) and simulated urban patterns (in Fig. \ref{fig:Fake_real}(b)). On a visual inspection, simulations are practically indistinguishable from the actual urban patterns, with realistic densities and complexity of settlement patterns. Input images and generated images are exhibiting realistic concentration at the center and distribution of settlement in the surrounding regions. Various quantitative metrics as discussed earlier are used to evaluate the performance of the Indian CityGAN model \cite{bertaud2003spatial}.

\begin{figure}[ht]
\centering
\includegraphics[width=\linewidth]{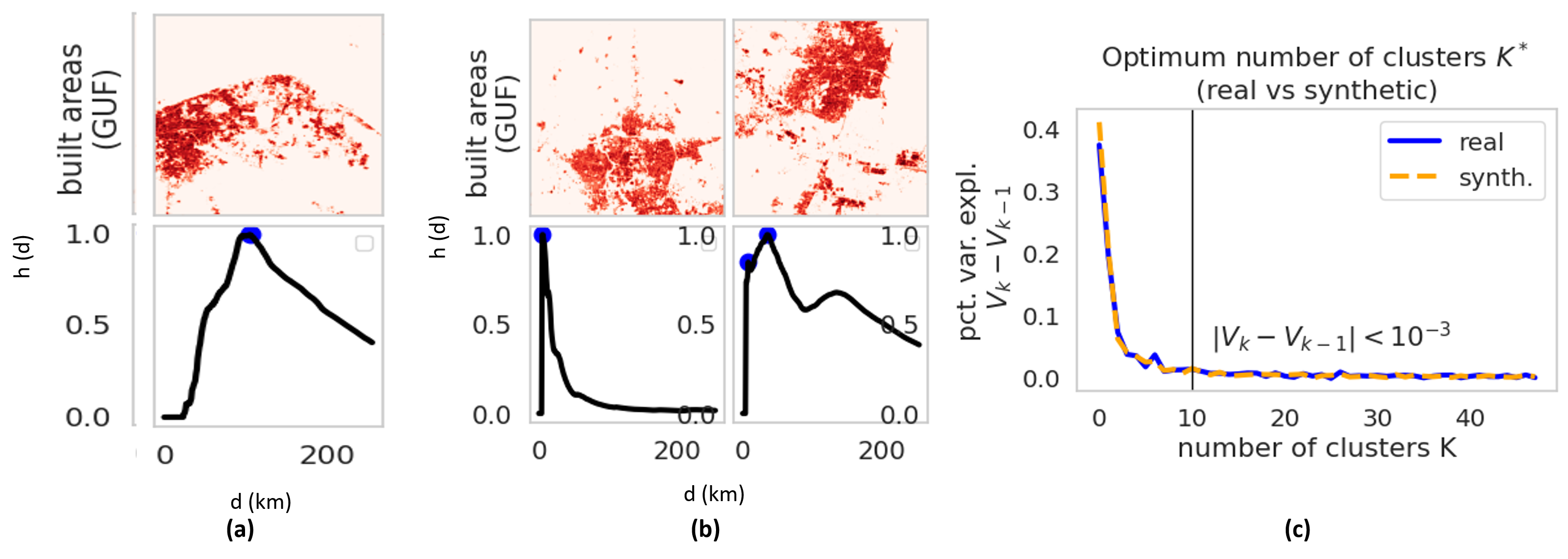}
\caption{(a) Average radial profile map of Bhagalpur city; (b) Human settlement maps of two example cities (top row) along with their average radial profiles (bottom row), where the blue dots represent the peaks found by the peak-search algorithm; (c) Clustered radial profiles of real and synthetic settlement patterns for determining the number of clusters.}
\label{fig:APR_cluster}
\end{figure}

Among various spatial statistical measures, ARP \cite{Albert2018ModelingUP} is used to compare the real and simulated urban patterns. We utilize Eq. \ref{eqn:h(d)} to compile the polycentric nature of real and generated images via the peak search algorithm illustrated in Fig. \ref{fig:Bihar}. 
The peak search algorithm finds points in univariate profiles whose value (peak height) is a fraction of the maximum height $h$ and at a distance from the previously identified peak of at least $d$. We set an acceptable value of $h=80\%$  and  $d = 430 m$ via the cross-validation method. Graphical representations of peak search outcomes are illustrated in Figs. \ref{fig:APR_cluster} (a) and (b). 

\begin{figure}[ht]
\centering
\includegraphics[width=\linewidth]{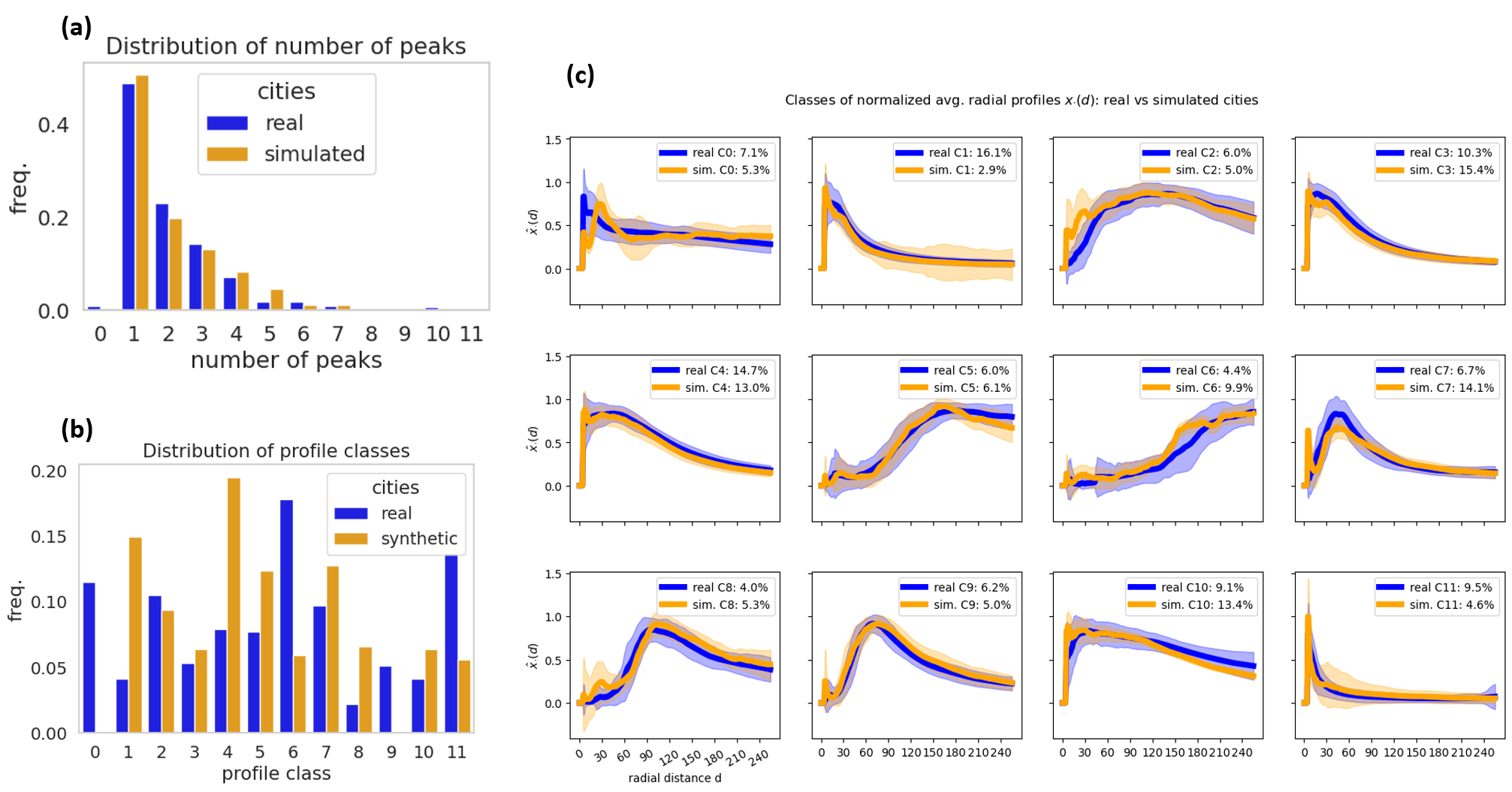}
\caption{(a) Comparison of the distribution of the number of peaks of real and generated cities; (b) Comparison of the distribution of average radial profile classes of real and generated cities; (c) The typical radial profiles for real and generated Indian cities (similar profile)}
\label{fig:GAN_validation}
\end{figure}

The distribution of the number of peaks for real and fake cities is compared (refer to Fig. \ref{fig:GAN_validation}(a)) and similarities between the distributions are found. Further, we cluster the radial profiles of real cities using K-means algorithm \cite{ahmed2020k, Hastie2005TheEO} and compare it to the typical profile of generated cities. Fig. \ref{fig:GAN_validation} presents a summary of our findings from the CityGAN model. Our analysis suggests that $K^*=10$ (refer Fig. \ref{fig:APR_cluster}(c)) gives the optimal number of clusters for both actual and synthetic scenes using a straightforward fraction of the sum-of-squares argument \cite{Hastie2005TheEO}. In Fig. \ref{fig:GAN_validation}(c), the distribution of scenes by class is given and they have a similar shape and are more comparable. But in Fig. \ref{fig:GAN_validation}(b), we find more disparities for classes 1 and 4 (monocentric), 6, 8, and 11 (sprawled patterns). These discrepancies may result from a sampling technique that would have favored the abundance of mono-centric urban patterns while the simulation was produced regardless of the location of the urban center. Experimental results show that using the WSF dataset, CityGAN generates precise urban patterns for Indian cities. 

\subsection*{Relationship between human settlement and transportation network of real cities}

\begin{figure*}[h]
	\centering
	\includegraphics[width=18cm, height=8cm]{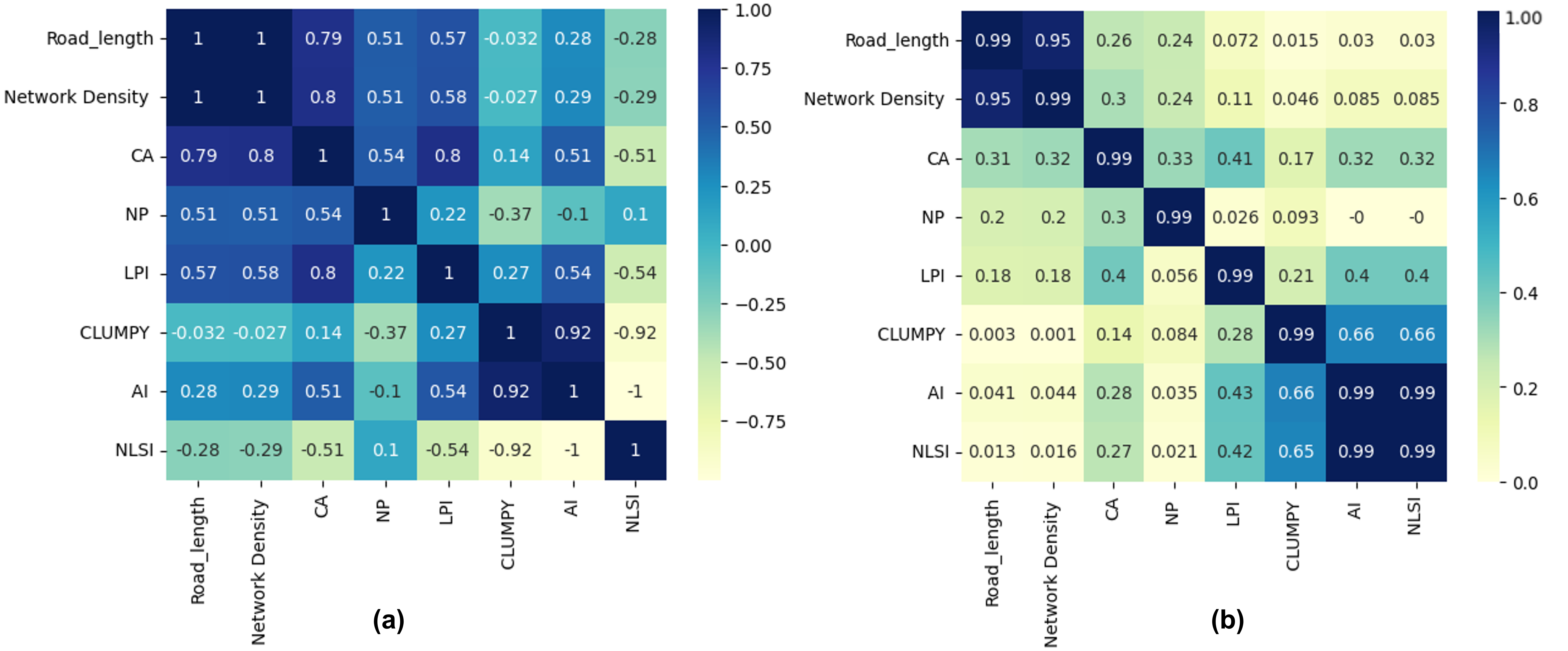}
	\caption{Heat map of the correlation between transportation indices and human settlement indices, the (a) PCC, and (b) CCC based on the input data.}
 \label{fig:pcc_ccc}
\end{figure*}

HSIs and TI are computed for the selected cities (workflow is illustrated in Figs. \ref{fig:methods} (b) and (c)). The outcomes of the analysis of human settlements using selected spatial indices are displayed in Table \ref{tab:HSI}. Table \ref{tab:TI} provides examples of the calculation of the transportation index (network density) and Table \ref{tab:RL} shows descriptive statistics of network density. The result shows the spatial distributions of network density vary among cities. Once the metrics are derived, correlation coefficients (CC) are calculated to determine the relationship between the human settlement indices (CA, NP, CLUMPY, LPI, AI, and NLSI) and the transportation index (RL, ND). The heat maps of the correlations between transportation indices and human settlement indices are illustrated in Fig. \ref{fig:pcc_ccc}.  

\begin{table*}
\centering

\resizebox{\textwidth}{!}{%
%\newpage
%\begin{center}
%\setlength{\tabcolsep}{1pt}
\begin{tabular}{|l  l  l  l  l  l  l  l| p{0.1\linewidth}|}
 \hline
 Grid number & City name & CA & NP & LPI & CLUMPY & AI & NLSI\\ [0.5ex] 
 \hline
 G1 & New Delhi & 4054.11 & 5299 & 3.782 & 0.680 & 79.607 & 0.203
 \\
 \hline
 G2 & Belgaum & 1534.05 & 5999 & 0.234 & 0.637 & 68.421 & 0.315
\\
\hline
G3 & Gulbarga & 2377.14 & 5722 & 0.844 & 0.652 & 72.184 & 0.278
\\
\hline
G4 & Jamnagar & 2141.93 & 3003 & 1.857 & 0.733 & 78.272 & 0.217
\\
\hline
G5 & Dhulia & 1608.7 & 4145 & 0.629 & 0.692 & 73.565 & 0.264
\\
\hline
\end{tabular}
}
\caption{Human settlement indices of small and medium-sized towns/cities in India calculated for the year 2019}
\label{tab:HSI}
\end{table*}

\begin{table}
\centering

%\begin{center}
\setlength{\tabcolsep}{4pt}
% \resizebox{\textwidth}{!}
{%
\begin{tabular}{|c c c c c c |} 
 \hline
 {SI No} & {City name} & {n} & {L (km)} & {A} (sq. km) & {ND} \\ 
 \hline
 1 & New Delhi & G1 & 1898.33 & 108.4 & 17.513 \\
 \hline
 2 & Belgaum & G2 & 1313.71 & 108.4 & 11.084 \\
 \hline
 3 & Gulbarga & G3 & 1714.51 & 108.4 & 14.574 \\
 \hline
 4 & Jamnagar & G4 & 1454.75 & 108.4 & 12.764 \\
 \hline
 5 & Dhulia & G5 & 1164.13 & 108.4 & 10.106 \\
 \hline
\end{tabular}
}
\caption{City number (n), the total length of the road network (L), City buffer area (A) and network density (ND)}

\label{tab:TI}
%\end{center}
\end{table}

\begin{table}
\centering
%\begin{center}
\setlength{\tabcolsep}{4pt}
\begin{tabular}{|c c c c c|} 
 \hline
 {Transportation metrics} & {Min} & {Max} & {Mean} & {std. deviation} \\ [0.5ex] 
 \hline
 RL & 15.7 & 2791.51 & 434.514 & 392.389\\
 \hline
\end{tabular}
\caption{Descriptive statistics of transportation metrics (namely road length)}

\label{tab:RL}
\end{table}

Fig. \ref{fig:pcc_ccc}(a) demonstrates the value matrix of Pearson’s correlation coefficient (PCC) as a measure of a linear relationship (ranges between $-1$ to $1$). Its value reflects the strength of the link between metrics. Positive numbers demonstrate the beneficial influence of variables on each other, and negative values represent the negative connection between variables. Here, the correlation index is displayed by the color intensity as well. The correlation increases as the color bar rises, light yellow color denotes a lower correlation. As shown in Fig. \ref{fig:pcc_ccc}(a), the PCC values of coverage measures RL and ND are highly related to each other; therefore we choose ND as the response variable of TI. Settlement metrics such as CLUMPY and NLSI have the minimum correlation with transportation metrics (light blue color), while CA demonstrates the maximum value of correlation coefficient with transportation variables. The correlation coefficients of RL and ND with CA are 0.7 and 0.80 respectively. According to PCC, the highest correlation exists between CA and RL, CA and ND. Hence, CA is an inevitable variable in our regression model. Nonlinear relationships between HSIs and TIs are explored using Chatterjee correlation coefficients (CCC) and reported in Fig. \ref{fig:pcc_ccc}(b). In the case of CCC, values reflect that most variables have positive correlations and CA has the highest rank followed by NP and LPI. 

\subsection*{Prediction of transportation index}

A supervised machine learning approach (Kernel Ridge Regression) was implemented to predict road network density for given settlement patterns (refer to Fig. \ref{fig:methods} (d)). To train and evaluate the prediction model, the raw dataset (database of 503 real cities) was divided into two parts: training (80\%) and testing (20\%). We compare the performance of various other supervised regression models such as support vector regression (SVR), decision tree (DT), gradient boosting (GB), multilayer perceptron regression (MLP), XGboost (XGB), linear regression (LR), random forest (RF) and simple ridge regression (RR) to select the best-fit model. To validate the models, four statistical scores: Mean Squared Error (MSE), Mean Absolute Error (MAE), R-Squared and Adjusted R-Squared were used. The results of our model and comparison with other models are summarized in Table \ref{tab:validation metrics}. Experimental result shows that KRR outperforms all other eight state-of-the-art regression models, in terms of the highest $R^2$ and Adjusted $R^2$ and lowest error metrics (MSE and MAE values). Because of the ability to handle nonlinearity and multicollinearity difficulties within datasets, the KRR regression model performs best. Validation metrics indicate that our model is good at predicting network density for urban patterns. This implies that the supervised KRR model can be applied to predict TI for newly generated cities by CityGAN. 
\begin{table*}
\centering

%\begin{center}
%\setlength{\tabcolsep}{4pt}
\resizebox{\textwidth}{!}{%
\begin{tabular}{|c  c c  c c  c  c  c  c  c |} 
 \hline
 {Accuracy metrics} & {SVR}  & {DT} & {GB} & MLP & {XGB} & {LR} & {RF} & {RR} &  {KRR}\\ [0.5ex] 
 \hline
 MSE & 6.439 & 5.792 & 5.265 & 4.505 & 4.448 & 4.343 & 4.133 & 3.998 & \textbf{3.661}
 \\
 \hline
 MAE & 1.832 & 1.716 & 1.688 & 1.582 & 1.535 & 1.576 & 1.475 & 1.511 & \textbf{1.397}
 \\
 \hline
 $R^2$ Score & 0.490 & 0.541 & 0.583 & 0.643 & 0.648 & 0.656 & 0.673 & 0.683 & \textbf{0.710}
\\
\hline
Adj $R^2$ Score & 0.463 & 0.517 & 0.561 & 0.624 & 0.629 & 0.638 & 0.655 & 0.667 & \textbf{0.695}
% \hline
 \\
 \hline
\end{tabular}
}
\caption{Performance metrics for the test set of real dataset. Best model's values are highlighted in \textbf{bold}.}
\label{tab:validation metrics}
%\end{center}
\end{table*}

\section*{Discussion}
India is now the most populous country in the world \cite{united2022world} with more than 30\% of the population residing in the urban area. Government of India recently came up with a scheme called IDSMT \cite{ganguly201614} to improve urban planning and road networks for the development of small and medium-sized cities (population size up to 500k). The Ministry of Urban Development offers financial and technical assistance to local bodies in developing their region's infrastructure and basic services to promote inclusive growth and balanced regional development. One of the objectives of this scheme is to address the transportation infrastructural deficits and service shortfalls of the eligible cities. However, Central Government finds difficulty in making a correct decision to allocate funds for transportation infrastructure development. 

This study proposes a hybrid model to predict road network density for small and medium-sized settlement patterns in India. We used the publicly available WSF datasets and CityGAN model to build an unsupervised model to simulate realistic Indian urban patterns. The average radial profile was used to compare real against simulated cities to validate the performance of CityGAN. Also, we used the K-means technique to cluster the radial profiles with the optimal number of clusters to be equal to 10 for both actual and synthetic scenes using a straightforward fraction of sum-of-squares reasoning. Landscape structures of these generated cities were measured in terms of human settlement indices using spatial landscape metrics. Then, various supervised machine learning models were implemented for predicting transportation indices from human settlement indices based on real city datasets. All the regression models were compared based on error measurement metrics. The performance of the Kernel Ridge Regression (KRR) model outperformed the benchmark regression methods. The transportation index estimated from the KRR model is compared with actual test data (from real imagery). KRR has comparatively less MAE of 1.39 and a higher $R^2$ value of 0.71. Thus, the proposed hybrid model can be used to predict the transportation index in terms of road network density for CityGAN-generated towns and cities. Our proposal can be treated as a versatile decision-support system for sustainable planning and development of new small and medium-sized towns and cities in terms of transportation infrastructure. However, the proposed method is useful to establish a relationship between coverage measures of transportation variables with HSIs, but the current work doesn't consider any connectivity measure. As a future scope of work, making interconnections between connectivity measures with HSIs will be worth exploring.

% Up to three levels of \textbf{subheading} are permitted. Subheadings should not be numbered.

% \subsection*{Subsection}

% Example text under a subsection. Bulleted lists may be used where appropriate, e.g.

% \begin{itemize}
% \item First item
% \item Second item
% \end{itemize}

% \subsubsection*{Third-level section}
 
% Topical subheadings are allowed.

% \section*{Discussion}

% The Discussion should be succinct and must not contain subheadings.

\section*{Methods}

In this section, we introduce the components of the model used in the hybrid framework are demonstrated. First, we discuss various correlation measures that are used in this study. Further, our proposed two-step pipeline approach utilizes the popularly used GAN model (CityGAN) in this case and the KRR model, a nonlinear shrinkage method for regression modeling. After this, we go over the suggested RidgeGAN method.

\subsection*{Correlation analysis}
Correlation coefficients (CC) are popular statistical measures used to determine the strength of a relationship between two or more variables (can be numerical or categorical). Pearson's correlation coefficient (PCC) is the most commonly used classical method of measuring linear associations, and its ease of use is advantageous \cite{Sadeghi2022ChatterjeeCC}. However, their efficiency may be limited when dealing with non-normal, noisy, closed, or open data (even after applying log ratios to the data). Chatterjee Correlation Coefficient (CCC) \cite{Chatterjee2019ANC}, a recently developed method based on cross-correlation between ranked increments, is a reliable alternative to traditional correlation methods. CCC can deal with data that contains outliers or has non-normal distributions and it does not make any assumptions about the data distribution \cite{Sadeghi2022ChatterjeeCC}. A Python implementation of CCC can be done using the ``TripleCpy'' package in Python. We can define CCC mathematically as follows: Given a pair of random variables $(X, Y)$ and suppose the realizations $X_i$'s and $Y_i$'s have no tie. A rearrangement of the pairs as $\left(X_{(1)}, Y_{(1)}\right), \ldots,\left(X_{(n)}, Y_{(n)}\right)$ is done such that $X_{(1)} \leq \cdots \leq X_{(n)}$. Let $r_i$ be the rank of $Y_{(i)}$ then CCC is defined using the formula:
$$
\xi_n(X, Y):=1-\frac{3 \sum_{i=1}^{n-1}\left|r_{i+1}-r_i\right|}{n^2-1},
$$
where $n$ is the number of observations. Once the relationships are assessed between TI and HSIs, selecting the ``best'' regression model is easier so that the model explains the variability in the response variables with possibly lower prediction error. However, the ridge regression (RR) method can directly handle multicollinearity structures in the data along with the instability of least square estimators \cite{Hoerl2000RidgeRB}. A more effective nonlinear regularized regression technique in machine learning is kernel ridge regression (KRR). The creation of the ridge regression method addresses some of the shortcomings of the least square method (over-fitting and multicollinearity) \cite{Hoerl1970RidgeRA}. One of their advantages is that the kernel implementation allows to handle nonlinearity of the data \cite{Maalouf2011KernelLR, Exterkate2011ModellingII, Saunders1998RidgeRL}.

\subsection*{CityGAN: Generative Adversarial Networks for modeling urban patterns}

GAN is a powerful unsupervised deep learning model that learns representations of input data to fit high dimensional complex distributions \cite{Goodfellow2014GenerativeAN}. GANs are revolutionary in that they can produce very high-quality (i.e., extremely realistic) samples compared to predecessor models at similar computational costs. In general, GAN is a system that consists of two neural networks competing against each other in a zero-sum game context \cite{goodfellow2016deep}. The two neural network architectures are a generator ($G$) and a discriminator ($D$) and they can generate new data that conforms to learned patterns through both generative and adversarial processes. GANs demonstrate promising performance in modeling complex geospatial data having spatial dependence \cite{Wu2022GenerativeAN}. Whilst the core application of GANs has been computer vision and image processing \cite{Goodfellow2014GenerativeAN}; however, their use in geoscience has provided urban planners with novel ways of generating ``new'' samples that can easily outperform state-of-the-art geostatistical tools. In this study, we deploy CityGAN \cite{Albert2018ModelingUP} to learn the urban patterns from real settlement images and generate hyper-realistic urban settlement images.  $G$ and $D$ are both deep convolutional neural networks with weight vectors  $\theta_G$ and $\theta_D$. Back-propagation is used to learn these weights by alternatively reducing the following loss functions
\begin{equation}
    \theta_D: \mathcal{L}_D=E_{h \sim p_h}[\log D(H)]+E_{z \sim p_z}[\log (1-D(G(z)))]
    \label{eqn:1}
\end{equation}

\begin{equation}
    \theta_G: \mathcal{L}_G = E_{z \sim p_z}[\log (1-D(G(z)))]
    \label{eqn:2}
\end{equation}

Here, the generator is made up of numerous convolutional blocks, including inverse-convolutional, batch normalization, and rectified linear unit (ReLU) layers, and ends with a hyperbolic tangent layer (which applies tanh ($\cdot$) nonlinearity to each element of the produced map). Recent modifications of GANs have allowed performing conditional generation as domain transformation. In the GAN training phase, it is worth noting that the generator network is usually able to create realistic samples whereas the discriminator is an auxiliary network that gets discarded after training. Once GAN is trained, the CityGAN \cite{Albert2018ModelingUP} can be used to generate new synthetic urban images that can be used for a variety of applications such as urban planning, disaster response, and simulations. Iteratively optimizing the $G$ and $D$ networks is part of the training process. The generator network attempts to deceive the discriminator by generating images that resemble real urban images, whereas the discriminator network attempts to correctly classify whether an image is real or fake. The networks are updated based on classification and generation errors until the generator produces images that are indistinguishable from real-world human settlement scenes. $H_{fake}$ is implicitly sampled from the data distribution that the generator tries to imitate when G is at its optimum. However, the GAN-generated images may not be representative of all possible urbanization patterns because they are based on the training dataset and the GAN architecture used. As a result, before using GAN-generated images for any practical application, it is critical to carefully evaluate them and compare them to real-world urban areas.

\subsection*{Kernel Ridge Regression (KRR)}
Regression modeling is a fundamental area of machine learning where the target variable is quantitative (real numbers) in nature. The most classical approach is linear regression using the ordinary least square method. However, it has salient disadvantages, e.g., overfitting and multicollinearity which can be addressed via ridge regression. Ridge regression ``shrinks'' the least square coefficients using regularization parameter via minimizing the objective function given below:
\begin{equation}
    \text{Ridge}(\beta) = \frac{1}{2}(Y-X\beta)^{T}(Y-X\beta) +\frac{\lambda}{2}\beta^{T}\beta,
    \label{eqn:3}
\end{equation}
where $X \in R^{N \times D}$ is the feature matrix with $N$ being the number of training samples, $D$ is the number of features $Y\in R^1$ is the real-valued target vector, $\beta$ is the regression coefficients, and $\lambda \geq 0$ is the regularizer that helps in dealing multicollinearity problem. However, the Ridge regression model still has troubles when dealing with nonlinear data data~\cite{maalouf2011kernel}. A more general framework can be achieved by using a nonlinear mapping function $\phi(\cdot)$ that maps low dimension feature to high dimension (helps in learning nonlinear patterns). Now, a kernel function in the form of the dot product is used to avoid the cause of dimensionality of the nonlinear transformation. Mathematically, Kernel between two points, say $x_m$ and $x_n$ is given by
\begin{equation}
    K(x_m,x_n)= \phi(x_m)^T\phi(x_n),
    \label{eqn:4}
\end{equation}
 which satisfies Mercer's condition~\cite{shawe2004kernel}. The major impact of Kernel is ridge regression that allows the identification of nonlinear functional relationships between one variable with remaining features. In this study, we use radial basis kernel function (RBF)~\cite{Saunders1998RidgeRL} which is defined by:
\begin{equation}
    K(x_m,x_n)= e^{-\gamma \| x_m - x_n\|^2},
    \label{eqn:5}
\end{equation}
where $\gamma$ is the width of the kernel. Predictions in KRR model for a new test input $x_{*}$ is given by,
\begin{equation}
    \beta^{T}\phi(x_{*}) = \sum_{n=1}^{N} (XX^{T}+\lambda I_{N})^{-1}Y K(x_n,x_*).
    \label{eqn:6}
\end{equation}
We use KRR to establish the relationship between HSIs and TI as shown in Fig. \ref{fig:methods} (d).

\subsection*{Hybrid model: RidgeGAN}

RidgeGAN is a hybrid approach based on unsupervised CityGAN and supervised KRR models. KRR \cite{Saunders1998RidgeRL} has a built-in mechanism to perform nonlinear regularization analysis in the presence of multicollinearity. CityGAN \cite{Albert2018ModelingUP} became popular to generate fake city images (a.k.a possible future cities) that look like real cities from the visual inspection and are statistically significant via learning the urban morphology. After implementation, we evaluated the performance of CityGAN, by comparing the real and simulated cities using the most widely used spatial summary statistics in an urban analysis called average radial profile and peak search algorithm. Each city is represented as a $10.5\times10.5$ km image covering the urban center and surrounding regions. Although, the quantifying transportation index has a vital role in the development of sustainable city planning and management. Here, we built a supervised KRR model to predict the transportation index by learning the relationship between urban patterns and the road transportation index. The KRR prediction model is integrated with the CityGAN model to predict the transportation index of newly generated cities using CityGAN. To build our hybrid model, we mainly use two models: an unsupervised learning model for generating urban patterns and a supervised learning model to predict the transportation index. To sum up, the workflow of the proposed RidgeGAN is detailed as follows (also see Fig. \ref{fig:methods} for a schematic workflow):
\begin{itemize}
  \item First, we apply CityGAN, an unsupervised learning model to generate small and medium-sized Indian cities using the available urban morphological features.
  \item Landscape structures of real and generated cities are measured in terms of Human Settlement Indices (HSI) using spatial landscape metrics.  
  \item We assess the relations between two important features of urban forms (human settlement and transportation system) and build a KRR model to predict the transportation index, namely network density. 
  \item The proposed hybrid model framework can predict the road network density on a given urban pattern for the urban universes generated in the first step. 
\end{itemize}

\bibliography{sample}

\begin{thebibliography}{10}
\urlstyle{rm}
\expandafter\ifx\csname url\endcsname\relax
  \def\url#1{\texttt{#1}}\fi
\expandafter\ifx\csname urlprefix\endcsname\relax\def\urlprefix{URL }\fi
\expandafter\ifx\csname doiprefix\endcsname\relax\def\doiprefix{DOI: }\fi
\providecommand{\bibinfo}[2]{#2}
\providecommand{\eprint}[2][]{\url{#2}}

\bibitem{lv2021land}
\bibinfo{author}{Lv, Z.}, \bibinfo{author}{Liu, T.},
  \bibinfo{author}{Benediktsson, J.~A.} \& \bibinfo{author}{Falco, N.}
\newblock \bibinfo{journal}{\bibinfo{title}{Land cover change detection
  techniques: Very-high-resolution optical images: A review}}.
\newblock {\emph{\JournalTitle{IEEE Geoscience and Remote Sensing Magazine}}}
  \textbf{\bibinfo{volume}{10}}, \bibinfo{pages}{44--63}
  (\bibinfo{year}{2021}).

\bibitem{li2021understanding}
\bibinfo{author}{Li, Z.}, \bibinfo{author}{Jiao, L.}, \bibinfo{author}{Zhang,
  B.}, \bibinfo{author}{Xu, G.} \& \bibinfo{author}{Liu, J.}
\newblock \bibinfo{journal}{\bibinfo{title}{Understanding the pattern and
  mechanism of spatial concentration of urban land use, population and economic
  activities: A case study in wuhan, china}}.
\newblock {\emph{\JournalTitle{Geo-spatial Information Science}}}
  \textbf{\bibinfo{volume}{24}}, \bibinfo{pages}{678--694}
  (\bibinfo{year}{2021}).

\bibitem{mahmoudzadeh2022urban}
\bibinfo{author}{Mahmoudzadeh, H.}, \bibinfo{author}{Abedini, A.} \&
  \bibinfo{author}{Aram, F.}
\newblock \bibinfo{journal}{\bibinfo{title}{Urban growth modeling and
  land-use/land-cover change analysis in a metropolitan area (case study:
  Tabriz)}}.
\newblock {\emph{\JournalTitle{Land}}} \textbf{\bibinfo{volume}{11}},
  \bibinfo{pages}{2162} (\bibinfo{year}{2022}).

\bibitem{rana2021prediction}
\bibinfo{author}{Rana, M.~S.} \& \bibinfo{author}{Sarkar, S.}
\newblock \bibinfo{journal}{\bibinfo{title}{Prediction of urban expansion by
  using land cover change detection approach}}.
\newblock {\emph{\JournalTitle{Heliyon}}} \textbf{\bibinfo{volume}{7}},
  \bibinfo{pages}{e08437} (\bibinfo{year}{2021}).

\bibitem{vani2020assessment}
\bibinfo{author}{Vani, M.} \& \bibinfo{author}{Prasad, P. R.~C.}
\newblock \bibinfo{journal}{\bibinfo{title}{Assessment of spatio-temporal
  changes in land use and land cover, urban sprawl, and land surface
  temperature in and around vijayawada city, india}}.
\newblock {\emph{\JournalTitle{Environment, Development and Sustainability}}}
  \textbf{\bibinfo{volume}{22}}, \bibinfo{pages}{3079--3095}
  (\bibinfo{year}{2020}).

\bibitem{sanga2022top}
\bibinfo{author}{Sanga, N.}, \bibinfo{author}{Gonzalez~Benson, O.} \&
  \bibinfo{author}{Josyula, L.}
\newblock \bibinfo{journal}{\bibinfo{title}{Top-down processes derail bottom-up
  objectives: a study in community engagement and ‘slum-free city
  planning’}}.
\newblock {\emph{\JournalTitle{Community Development Journal}}}
  \textbf{\bibinfo{volume}{57}}, \bibinfo{pages}{615--634}
  (\bibinfo{year}{2022}).

\bibitem{falah2020urban}
\bibinfo{author}{Falah, N.}, \bibinfo{author}{Karimi, A.} \&
  \bibinfo{author}{Harandi, A.~T.}
\newblock \bibinfo{journal}{\bibinfo{title}{Urban growth modeling using
  cellular automata model and ahp (case study: Qazvin city)}}.
\newblock {\emph{\JournalTitle{Modeling Earth Systems and Environment}}}
  \textbf{\bibinfo{volume}{6}}, \bibinfo{pages}{235--248}
  (\bibinfo{year}{2020}).

\bibitem{mathioulakis2017using}
\bibinfo{author}{Mathioulakis, S.} \& \bibinfo{author}{Photis, Y.~N.}
\newblock \bibinfo{journal}{\bibinfo{title}{Using the sleuth model to simulate
  future urban growth in the greater eastern attica area, greece}}.
\newblock {\emph{\JournalTitle{European Journal of Geography}}}
  \textbf{\bibinfo{volume}{8}} (\bibinfo{year}{2017}).

\bibitem{hamdy2016applying}
\bibinfo{author}{Hamdy, O.}, \bibinfo{author}{Zhao, S.},
  \bibinfo{author}{Osman, T.}, \bibinfo{author}{Salheen, M.~A.} \&
  \bibinfo{author}{Eid, Y.~Y.}
\newblock \bibinfo{journal}{\bibinfo{title}{Applying a hybrid model of markov
  chain and logistic regression to identify future urban sprawl in abouelreesh,
  aswan: A case study}}.
\newblock {\emph{\JournalTitle{Geosciences}}} \textbf{\bibinfo{volume}{6}},
  \bibinfo{pages}{43} (\bibinfo{year}{2016}).

\bibitem{liang2021understanding}
\bibinfo{author}{Liang, X.} \emph{et~al.}
\newblock \bibinfo{journal}{\bibinfo{title}{Understanding the drivers of
  sustainable land expansion using a patch-generating land use simulation
  (plus) model: A case study in wuhan, china}}.
\newblock {\emph{\JournalTitle{Computers, Environment and Urban Systems}}}
  \textbf{\bibinfo{volume}{85}}, \bibinfo{pages}{101569}
  (\bibinfo{year}{2021}).

\bibitem{pravitasari2018identifying}
\bibinfo{author}{Pravitasari, A.} \emph{et~al.}
\newblock \bibinfo{title}{Identifying the driving forces of urban expansion and
  its environmental impact in jakarta-bandung mega urban region}.
\newblock In \emph{\bibinfo{booktitle}{IOP conference series: earth and
  environmental science}}, vol. \bibinfo{volume}{149}, \bibinfo{pages}{012044}
  (\bibinfo{organization}{IOP Publishing}, \bibinfo{year}{2018}).

\bibitem{united2022world}
\bibinfo{author}{Nations, U.}
\newblock \bibinfo{title}{World population to reach 8 billion on 15 november
  2022} (\bibinfo{year}{2022}).

\bibitem{ritchie2018urbanization}
\bibinfo{author}{Ritchie, H.} \& \bibinfo{author}{Roser, M.}
\newblock \bibinfo{journal}{\bibinfo{title}{Urbanization}}.
\newblock {\emph{\JournalTitle{Our world in data}}}  (\bibinfo{year}{2018}).

\bibitem{ganguly1997integrated}
\bibinfo{author}{Ganguly, R.}
\newblock \bibinfo{journal}{\bibinfo{title}{Integrated development of small and
  medium towns in india}}.
\newblock {\emph{\JournalTitle{Regional science in developing countries}}}
  \bibinfo{pages}{196--211} (\bibinfo{year}{1997}).

\bibitem{ganguly201614}
\bibinfo{author}{Ganguly, R.}
\newblock \bibinfo{journal}{\bibinfo{title}{14 integrated development of}}.
\newblock {\emph{\JournalTitle{Regional Science in Developing Countries}}}
  \bibinfo{pages}{196} (\bibinfo{year}{2016}).

\bibitem{athar2022financing}
\bibinfo{author}{Athar, S.}, \bibinfo{author}{White, R.} \&
  \bibinfo{author}{Goyal, H.}
\newblock \emph{\bibinfo{title}{Financing India’s urban infrastructure
  needs}} (\bibinfo{publisher}{Washington, DC: World Bank},
  \bibinfo{year}{2022}).

\bibitem{engstrom2020government}
\bibinfo{author}{Engstrom, D.~F.}, \bibinfo{author}{Ho, D.~E.},
  \bibinfo{author}{Sharkey, C.~M.} \& \bibinfo{author}{Cu{\'e}llar, M.-F.}
\newblock \bibinfo{journal}{\bibinfo{title}{Government by algorithm: Artificial
  intelligence in federal administrative agencies}}.
\newblock {\emph{\JournalTitle{NYU School of Law, Public Law Research Paper}}}
  (\bibinfo{year}{2020}).

\bibitem{Goodfellow2014GenerativeAN}
\bibinfo{author}{Goodfellow, I.~J.} \emph{et~al.}
\newblock \bibinfo{title}{Generative adversarial nets}.
\newblock In \emph{\bibinfo{booktitle}{NIPS}} (\bibinfo{year}{2014}).

\bibitem{wang2021generative}
\bibinfo{author}{Wang, Z.}, \bibinfo{author}{She, Q.} \& \bibinfo{author}{Ward,
  T.~E.}
\newblock \bibinfo{journal}{\bibinfo{title}{Generative adversarial networks in
  computer vision: A survey and taxonomy}}.
\newblock {\emph{\JournalTitle{ACM Computing Surveys (CSUR)}}}
  \textbf{\bibinfo{volume}{54}}, \bibinfo{pages}{1--38} (\bibinfo{year}{2021}).

\bibitem{pan2019recent}
\bibinfo{author}{Pan, Z.} \emph{et~al.}
\newblock \bibinfo{journal}{\bibinfo{title}{Recent progress on generative
  adversarial networks (gans): A survey}}.
\newblock {\emph{\JournalTitle{IEEE access}}} \textbf{\bibinfo{volume}{7}},
  \bibinfo{pages}{36322--36333} (\bibinfo{year}{2019}).

\bibitem{zhu2020spatial}
\bibinfo{author}{Zhu, D.} \emph{et~al.}
\newblock \bibinfo{journal}{\bibinfo{title}{Spatial interpolation using
  conditional generative adversarial neural networks}}.
\newblock {\emph{\JournalTitle{International Journal of Geographical
  Information Science}}} \textbf{\bibinfo{volume}{34}},
  \bibinfo{pages}{735--758} (\bibinfo{year}{2020}).

\bibitem{Wu2022GenerativeAN}
\bibinfo{author}{Wu, A.~N.}, \bibinfo{author}{Stouffs, R.} \&
  \bibinfo{author}{Biljecki, F.}
\newblock \bibinfo{journal}{\bibinfo{title}{Generative adversarial networks in
  the built environment: A comprehensive review of the application of gans
  across data types and scales}}.
\newblock {\emph{\JournalTitle{Building and Environment}}}
  (\bibinfo{year}{2022}).

\bibitem{Albert2018ModelingUP}
\bibinfo{author}{Albert, A.}, \bibinfo{author}{Strano, E.},
  \bibinfo{author}{Kaur, J.} \& \bibinfo{author}{Gonz{\'a}lez, M.~C.}
\newblock \bibinfo{journal}{\bibinfo{title}{Modeling urbanization patterns with
  generative adversarial networks}}.
\newblock {\emph{\JournalTitle{IGARSS 2018 - 2018 IEEE International Geoscience
  and Remote Sensing Symposium}}} \bibinfo{pages}{2095--2098}
  (\bibinfo{year}{2018}).

\bibitem{zhang2022metrogan}
\bibinfo{author}{Zhang, W.}, \bibinfo{author}{Ma, Y.}, \bibinfo{author}{Zhu,
  D.}, \bibinfo{author}{Dong, L.} \& \bibinfo{author}{Liu, Y.}
\newblock \bibinfo{title}{Metrogan: Simulating urban morphology with generative
  adversarial network}.
\newblock In \emph{\bibinfo{booktitle}{Proceedings of the 28th ACM SIGKDD
  Conference on Knowledge Discovery and Data Mining}},
  \bibinfo{pages}{2482--2492} (\bibinfo{year}{2022}).

\bibitem{ronneberger2015unet}
\bibinfo{author}{Ronneberger, O.}, \bibinfo{author}{Fischer, P.} \&
  \bibinfo{author}{Brox, T.}
\newblock \bibinfo{title}{U-net: Convolutional networks for biomedical image
  segmentation}.
\newblock In \emph{\bibinfo{booktitle}{Medical Image Computing and
  Computer-Assisted Intervention--MICCAI 2015: 18th International Conference,
  Munich, Germany, October 5-9, 2015, Proceedings, Part III 18}},
  \bibinfo{pages}{234--241} (\bibinfo{organization}{Springer},
  \bibinfo{year}{2015}).

\bibitem{forman1981patches}
\bibinfo{author}{Forman, R.~T.} \& \bibinfo{author}{Godron, M.}
\newblock \bibinfo{journal}{\bibinfo{title}{Patches and structural components
  for a landscape ecology}}.
\newblock {\emph{\JournalTitle{BioScience}}} \textbf{\bibinfo{volume}{31}},
  \bibinfo{pages}{733--740} (\bibinfo{year}{1981}).

\bibitem{forman1986landscape}
\bibinfo{author}{Forman, R.} \& \bibinfo{author}{Godron, M.}
\newblock \bibinfo{journal}{\bibinfo{title}{Landscape ecology wiley \& sons}}.
\newblock {\emph{\JournalTitle{New York, NY, US}}}  (\bibinfo{year}{1986}).

\bibitem{gokyer2013understanding}
\bibinfo{author}{G{\"o}kyer, E.}
\newblock \bibinfo{title}{Understanding landscape structure using landscape
  metrics}.
\newblock In \emph{\bibinfo{booktitle}{Advances in landscape architecture}}
  (\bibinfo{publisher}{IntechOpen}, \bibinfo{year}{2013}).

\bibitem{mobaied2016importance}
\bibinfo{author}{Mobaied, S.}, \bibinfo{author}{Geoffroy, J.-J.} \&
  \bibinfo{author}{Machon, N.}
\newblock \bibinfo{journal}{\bibinfo{title}{The importance of spatiotemporal
  heterogeneity for biodiversity in forest—heathland mosaics and implications
  for heathland conservation}}.
\newblock {\emph{\JournalTitle{Journal of Environmental Protection}}}
  \textbf{\bibinfo{volume}{7}}, \bibinfo{pages}{1317--1332}
  (\bibinfo{year}{2016}).

\bibitem{Aithal2020UrbanGP}
\bibinfo{author}{Aithal, B.~H.} \emph{et~al.}
\newblock \bibinfo{title}{Urban growth patterns in india}
  (\bibinfo{year}{2020}).

\bibitem{Sudhira2012EffectOL}
\bibinfo{author}{Sudhira, H.~S.}, \bibinfo{author}{Shetty, P.~J.},
  \bibinfo{author}{Gowda, S.~V.} \& \bibinfo{author}{Gururaja, K.~V.}
\newblock \bibinfo{journal}{\bibinfo{title}{Effect of landscape metrics on
  varied spatial extents of bangalore, india}}.
\newblock {\emph{\JournalTitle{Asian Journal of Geoinformatics}}}
  \textbf{\bibinfo{volume}{12}} (\bibinfo{year}{2012}).

\bibitem{wilson2003development}
\bibinfo{author}{Wilson, E.~H.}, \bibinfo{author}{Hurd, J.~D.},
  \bibinfo{author}{Civco, D.~L.}, \bibinfo{author}{Prisloe, M.~P.} \&
  \bibinfo{author}{Arnold, C.}
\newblock \bibinfo{journal}{\bibinfo{title}{Development of a geospatial model
  to quantify, describe and map urban growth}}.
\newblock {\emph{\JournalTitle{Remote sensing of environment}}}
  \textbf{\bibinfo{volume}{86}}, \bibinfo{pages}{275--285}
  (\bibinfo{year}{2003}).

\bibitem{esch2017breaking}
\bibinfo{author}{Esch, T.} \emph{et~al.}
\newblock \bibinfo{journal}{\bibinfo{title}{Breaking new ground in mapping
  human settlements from space--the global urban footprint}}.
\newblock {\emph{\JournalTitle{ISPRS Journal of Photogrammetry and Remote
  Sensing}}} \textbf{\bibinfo{volume}{134}}, \bibinfo{pages}{30--42}
  (\bibinfo{year}{2017}).

\bibitem{esch2018global}
\bibinfo{author}{Esch, T.}, \bibinfo{author}{Heldens, W.} \&
  \bibinfo{author}{Hirner, A.}
\newblock \bibinfo{title}{The global urban footprint}.
\newblock In \emph{\bibinfo{booktitle}{Urban Remote Sensing}},
  \bibinfo{pages}{3--14} (\bibinfo{publisher}{CRC Press},
  \bibinfo{year}{2018}).

\bibitem{roberts2022principles}
\bibinfo{author}{Roberts, D.~A.}, \bibinfo{author}{Yaida, S.} \&
  \bibinfo{author}{Hanin, B.}
\newblock \emph{\bibinfo{title}{The principles of deep learning theory}}
  (\bibinfo{publisher}{Cambridge University Press Cambridge, MA, USA},
  \bibinfo{year}{2022}).

\bibitem{Hong2017HowGA}
\bibinfo{author}{Hong, Y.}, \bibinfo{author}{Hwang, U.}, \bibinfo{author}{Yoo,
  J.} \& \bibinfo{author}{Yoon, S.}
\newblock \bibinfo{journal}{\bibinfo{title}{How generative adversarial networks
  and their variants work}}.
\newblock {\emph{\JournalTitle{ACM Computing Surveys (CSUR)}}}
  \textbf{\bibinfo{volume}{52}}, \bibinfo{pages}{1 -- 43}
  (\bibinfo{year}{2017}).

\bibitem{Lin2016MARTAGU}
\bibinfo{author}{Lin, D.}, \bibinfo{author}{Fu, K.}, \bibinfo{author}{Wang,
  Y.}, \bibinfo{author}{Xu, G.} \& \bibinfo{author}{Sun, X.}
\newblock \bibinfo{journal}{\bibinfo{title}{Marta gans: Unsupervised
  representation learning for remote sensing image classification}}.
\newblock {\emph{\JournalTitle{IEEE Geoscience and Remote Sensing Letters}}}
  \textbf{\bibinfo{volume}{14}}, \bibinfo{pages}{2092--2096}
  (\bibinfo{year}{2016}).

\bibitem{Jetchev2016TextureSW}
\bibinfo{author}{Jetchev, N.}, \bibinfo{author}{Bergmann, U.~M.} \&
  \bibinfo{author}{Vollgraf, R.}
\newblock \bibinfo{journal}{\bibinfo{title}{Texture synthesis with spatial
  generative adversarial networks}}.
\newblock {\emph{\JournalTitle{ArXiv}}}
  \textbf{\bibinfo{volume}{abs/1611.08207}} (\bibinfo{year}{2016}).

\bibitem{albert2019spatial}
\bibinfo{author}{Albert, A.}, \bibinfo{author}{Kaur, J.},
  \bibinfo{author}{Strano, E.} \& \bibinfo{author}{Gonzalez, M.}
\newblock \bibinfo{journal}{\bibinfo{title}{Spatial sensitivity analysis for
  urban land use prediction with physics-constrained conditional generative
  adversarial networks}}.
\newblock {\emph{\JournalTitle{arXiv preprint arXiv:1907.09543}}}
  (\bibinfo{year}{2019}).

\bibitem{Xing2022UnsupervisedDA}
\bibinfo{author}{Xing, S.} \emph{et~al.}
\newblock \bibinfo{journal}{\bibinfo{title}{Unsupervised domain adaptation gan
  inversion for image editing}}.
\newblock {\emph{\JournalTitle{ArXiv}}}
  \textbf{\bibinfo{volume}{abs/2211.12123}} (\bibinfo{year}{2022}).

\bibitem{Wu2002KeyIA}
\bibinfo{author}{Wu, J.} \& \bibinfo{author}{Hobbs, R.~J.}
\newblock \bibinfo{journal}{\bibinfo{title}{Key issues and research priorities
  in landscape ecology: An idiosyncratic synthesis}}.
\newblock {\emph{\JournalTitle{Landscape Ecology}}}
  \textbf{\bibinfo{volume}{17}}, \bibinfo{pages}{355--365}
  (\bibinfo{year}{2002}).

\bibitem{bertaud2003spatial}
\bibinfo{author}{Bertaud, A.} \& \bibinfo{author}{Malpezzi, S.}
\newblock \bibinfo{journal}{\bibinfo{title}{The spatial distribution of
  population in 48 world cities: Implications for economies in transition}}.
\newblock {\emph{\JournalTitle{Center for urban land economics research,
  University of Wisconsin}}} \textbf{\bibinfo{volume}{32}},
  \bibinfo{pages}{54--55} (\bibinfo{year}{2003}).

\bibitem{Brown2012SocialLM}
\bibinfo{author}{Brown, G.} \& \bibinfo{author}{Reed, P.~C.}
\newblock \bibinfo{journal}{\bibinfo{title}{Social landscape metrics: Measures
  for understanding place values from public participation geographic
  information systems (ppgis)}}.
\newblock {\emph{\JournalTitle{Landscape Research}}}
  \textbf{\bibinfo{volume}{37}}, \bibinfo{pages}{73 -- 90}
  (\bibinfo{year}{2012}).

\bibitem{bhat2015spatial}
\bibinfo{author}{Bhat, V.}, \bibinfo{author}{Aithal, B.~H.} \&
  \bibinfo{author}{Ramachandra, T.}
\newblock \bibinfo{journal}{\bibinfo{title}{Spatial patterns of urban growth
  with globalisation in india’s silicon valley}}.
\newblock {\emph{\JournalTitle{Organized By Department of Civil Engineering,
  Indian Institute of Technology (Banaras Hindu University), Varanasi-221005
  Uttar Pradesh, India}}} \textbf{\bibinfo{volume}{98}} (\bibinfo{year}{2015}).

\bibitem{Zhong2022AdvancesII}
\bibinfo{author}{Zhong, M.} \emph{et~al.}
\newblock \bibinfo{journal}{\bibinfo{title}{Advances in integrated land use
  transport modeling}}.
\newblock {\emph{\JournalTitle{Advances in Transport Policy and Planning}}}
  (\bibinfo{year}{2022}).

\bibitem{Aljoufie2011UrbanGA}
\bibinfo{author}{Aljoufie, M.}, \bibinfo{author}{Zuidgeest, M. H.~P.},
  \bibinfo{author}{Brussel, M. J.~G.} \& \bibinfo{author}{van Maarseveen, M.}
\newblock \bibinfo{journal}{\bibinfo{title}{Urban growth and transport:
  Understanding the spatial temporal relationship}}.
\newblock {\emph{\JournalTitle{WIT Transactions on the Built Environment}}}
  \textbf{\bibinfo{volume}{116}}, \bibinfo{pages}{315--328}
  (\bibinfo{year}{2011}).

\bibitem{Allaw2019ARS}
\bibinfo{author}{Allaw, K.}, \bibinfo{author}{G{\'e}rard, J.~A.},
  \bibinfo{author}{Chehayeb, M.} \& \bibinfo{author}{Saliba, N.~B.}
\newblock \bibinfo{title}{A remote sensing approach to calculate population
  using roads network data in lebanon} (\bibinfo{year}{2019}).

\bibitem{zeroual2021predicting}
\bibinfo{author}{Zeroual, A.}, \bibinfo{author}{Harrou, F.} \&
  \bibinfo{author}{Sun, Y.}
\newblock \bibinfo{title}{Predicting road traffic density using a machine
  learning-driven approach}.
\newblock In \emph{\bibinfo{booktitle}{2021 International Conference on
  Electrical, Computer and Energy Technologies (ICECET)}},
  \bibinfo{pages}{1--6} (\bibinfo{organization}{IEEE}, \bibinfo{year}{2021}).

\bibitem{Budiarto2014RoadDP}
\bibinfo{author}{Budiarto, J.}, \bibinfo{author}{Sulistyo, S.},
  \bibinfo{author}{Mustika, I.~W.} \& \bibinfo{author}{Infantono, A.}
\newblock \bibinfo{journal}{\bibinfo{title}{Road density prediction: Updated
  methods of turning probabilities and highway capacities manual for achieving
  the best route}}.
\newblock {\emph{\JournalTitle{2014 International Conference on Electrical
  Engineering and Computer Science (ICEECS)}}} \bibinfo{pages}{168--173}
  (\bibinfo{year}{2014}).

\bibitem{Thottolil2021AssessmentOT}
\bibinfo{author}{Thottolil, R.} \& \bibinfo{author}{Kumar, U.}
\newblock \bibinfo{journal}{\bibinfo{title}{Assessment of topological pattern
  of road network: A case study of bangalore city}}.
\newblock {\emph{\JournalTitle{2021 IEEE International India Geoscience and
  Remote Sensing Symposium (InGARSS)}}} \bibinfo{pages}{246--249}
  (\bibinfo{year}{2021}).

\bibitem{sharma2021small}
\bibinfo{author}{Sharma, A.}
\newblock \bibinfo{title}{Small towns in asia and urban sustainability}.
\newblock In \emph{\bibinfo{booktitle}{The Palgrave Encyclopedia of Urban and
  Regional Futures}}, \bibinfo{pages}{1--6} (\bibinfo{publisher}{Springer},
  \bibinfo{year}{2021}).

\bibitem{McGarigal1995FRAGSTATSSP}
\bibinfo{author}{McGarigal, K.} \& \bibinfo{author}{Marks, B.~J.}
\newblock \bibinfo{title}{Fragstats: spatial pattern analysis program for
  quantifying landscape structure.} (\bibinfo{year}{1995}).

\bibitem{chakraborty2019hybrid}
\bibinfo{author}{Chakraborty, T.}, \bibinfo{author}{Chakraborty, A.~K.} \&
  \bibinfo{author}{Mansoor, Z.}
\newblock \bibinfo{journal}{\bibinfo{title}{A hybrid regression model for water
  quality prediction}}.
\newblock {\emph{\JournalTitle{Opsearch}}} \textbf{\bibinfo{volume}{56}},
  \bibinfo{pages}{1167--1178} (\bibinfo{year}{2019}).

\bibitem{ahmed2020k}
\bibinfo{author}{Ahmed, M.}, \bibinfo{author}{Seraj, R.} \&
  \bibinfo{author}{Islam, S. M.~S.}
\newblock \bibinfo{journal}{\bibinfo{title}{The k-means algorithm: A
  comprehensive survey and performance evaluation}}.
\newblock {\emph{\JournalTitle{Electronics}}} \textbf{\bibinfo{volume}{9}},
  \bibinfo{pages}{1295} (\bibinfo{year}{2020}).

\bibitem{Hastie2005TheEO}
\bibinfo{author}{Hastie, T.~J.}, \bibinfo{author}{Tibshirani, R.} \&
  \bibinfo{author}{Friedman, J.~H.}
\newblock \bibinfo{title}{The elements of statistical learning: Data mining,
  inference, and prediction, 2nd edition}.
\newblock In \emph{\bibinfo{booktitle}{Springer Series in Statistics}}
  (\bibinfo{year}{2005}).

\bibitem{Sadeghi2022ChatterjeeCC}
\bibinfo{author}{Sadeghi, B.}
\newblock \bibinfo{journal}{\bibinfo{title}{Chatterjee correlation coefficient:
  a robust alternative for classic correlation methods in geochemical studies-
  (including “triplecpy” python package)}}.
\newblock {\emph{\JournalTitle{Ore Geology Reviews}}}  (\bibinfo{year}{2022}).

\bibitem{Chatterjee2019ANC}
\bibinfo{author}{Chatterjee, S.}
\newblock \bibinfo{journal}{\bibinfo{title}{A new coefficient of correlation}}.
\newblock {\emph{\JournalTitle{Journal of the American Statistical
  Association}}} \textbf{\bibinfo{volume}{116}}, \bibinfo{pages}{2009 -- 2022}
  (\bibinfo{year}{2019}).

\bibitem{Hoerl2000RidgeRB}
\bibinfo{author}{Hoerl, A.~E.} \& \bibinfo{author}{Kennard, R.~W.}
\newblock \bibinfo{journal}{\bibinfo{title}{Ridge regression: Biased estimation
  for nonorthogonal problems}}.
\newblock {\emph{\JournalTitle{Technometrics}}} \textbf{\bibinfo{volume}{42}},
  \bibinfo{pages}{80 -- 86} (\bibinfo{year}{2000}).

\bibitem{Hoerl1970RidgeRA}
\bibinfo{author}{Hoerl, A.~E.} \& \bibinfo{author}{Kennard, R.~W.}
\newblock \bibinfo{journal}{\bibinfo{title}{Ridge regression: Applications to
  nonorthogonal problems}}.
\newblock {\emph{\JournalTitle{Technometrics}}} \textbf{\bibinfo{volume}{12}},
  \bibinfo{pages}{69--82} (\bibinfo{year}{1970}).

\bibitem{Maalouf2011KernelLR}
\bibinfo{author}{Maalouf, M.}, \bibinfo{author}{Trafalis, T.~B.} \&
  \bibinfo{author}{Adrianto, I.}
\newblock \bibinfo{journal}{\bibinfo{title}{Kernel logistic regression using
  truncated newton method}}.
\newblock {\emph{\JournalTitle{Computational Management Science}}}
  \textbf{\bibinfo{volume}{8}}, \bibinfo{pages}{415--428}
  (\bibinfo{year}{2011}).

\bibitem{Exterkate2011ModellingII}
\bibinfo{author}{Exterkate, P.}
\newblock \bibinfo{journal}{\bibinfo{title}{Modelling issues in kernel ridge
  regression}}.
\newblock {\emph{\JournalTitle{ERN: Estimation (Topic)}}}
  (\bibinfo{year}{2011}).

\bibitem{Saunders1998RidgeRL}
\bibinfo{author}{Saunders, C.}, \bibinfo{author}{Gammerman, A.} \&
  \bibinfo{author}{Vovk, V.}
\newblock \bibinfo{title}{Ridge regression learning algorithm in dual
  variables}.
\newblock In \emph{\bibinfo{booktitle}{International Conference on Machine
  Learning}} (\bibinfo{year}{1998}).

\bibitem{goodfellow2016deep}
\bibinfo{author}{Goodfellow, I.}, \bibinfo{author}{Bengio, Y.} \&
  \bibinfo{author}{Courville, A.}
\newblock \emph{\bibinfo{title}{Deep learning}} (\bibinfo{publisher}{MIT
  press}, \bibinfo{year}{2016}).

\bibitem{maalouf2011kernel}
\bibinfo{author}{Maalouf, M.}, \bibinfo{author}{Trafalis, T.~B.},
  \bibinfo{author}{Adrianto, I.} \emph{et~al.}
\newblock \bibinfo{journal}{\bibinfo{title}{Kernel logistic regression using
  truncated newton method}}.
\newblock {\emph{\JournalTitle{Computational management science}}}
  \textbf{\bibinfo{volume}{8}}, \bibinfo{pages}{415} (\bibinfo{year}{2011}).

\bibitem{shawe2004kernel}
\bibinfo{author}{Shawe-Taylor, J.}, \bibinfo{author}{Cristianini, N.}
  \emph{et~al.}
\newblock \emph{\bibinfo{title}{Kernel methods for pattern analysis}}
  (\bibinfo{publisher}{Cambridge university press}, \bibinfo{year}{2004}).

\end{thebibliography}

% \noindent LaTeX formats citations and references automatically using the bibliography records in your .bib file, which you can edit via the project menu. Use the cite command for an inline citation, e.g.  \cite{Hao:gidmaps:2014}.

% For data citations of datasets uploaded to e.g. \emph{figshare}, please use the \verb|howpublished| option in the bib entry to specify the platform and the link, as in the \verb|Hao:gidmaps:2014| example in the sample bibliography file.

\section*{Data and Code availability statement }
The datasets and the codes for implementing the methods are made available at \url{https://github.com/Rahisha-Thottolil/RidgeGAN}.

\section*{Author contributions}

R.T. conceived the experiment(s) and collected the data, T.C. build the framework and R.T. analyzed the results, and R.T., T.C., and U.K. wrote and revised the manuscript. 

\section*{Additional information}
The authors declare that they have no known competing financial interests or personal relationships that could have appeared to influence the work reported in this paper.

% To include, in this order: \textbf{Accession codes} (where applicable); \textbf{Competing interests} (mandatory statement). 

% The corresponding author is responsible for submitting a \href{http://www.nature.com/srep/policies/index.html#competing}{competing interests statement} on behalf of all authors of the paper. This statement must be included in the submitted article file.

% \begin{figure}[ht]
% \centering
% \includegraphics[width=\linewidth]{stream}
% \caption{Legend (350 words max). Example legend text.}
% \label{fig:stream}
% \end{figure}

% \begin{table}[ht]
% \centering
% \begin{tabular}{|l|l|l|}
% \hline
% Condition & n & p \\
% \hline
% A & 5 & 0.1 \\
% \hline
% B & 10 & 0.01 \\
% \hline
% \end{tabular}
% \caption{\label{tab:example}Legend (350 words max). Example legend text.}
% \end{table}

% Figures and tables can be referenced in LaTeX using the ref command, e.g. Fig. \ref{fig:stream} and Table \ref{tab:example}.

\end{document}